\PassOptionsToPackage{table}{xcolor}
\documentclass[sigconf]{acmart}
\usepackage{amssymb}
\usepackage{bm}
\usepackage{mathtools}
\usepackage{makecell}
\usepackage{multirow}
\usepackage{listings}
\usepackage{pifont}

\definecolor{our_blue}{rgb}{0.85, 0.9, 0.95}
\definecolor{clip_red}{rgb}{0.961, 0.749, 0.737}  

\AtBeginDocument{%
  }

\copyrightyear{2025}
\acmYear{2025}
\setcopyright{acmlicensed}\acmConference[MM '25]{Proceedings of the 33rd
ACM International Conference on Multimedia}{October 27--31, 2025}{Dublin,
Ireland}
\acmBooktitle{Proceedings of the 33rd ACM International Conference on
Multimedia (MM '25), October 27--31, 2025, Dublin, Ireland}
\acmDOI{10.1145/3746027.3755137}
\acmISBN{979-8-4007-2035-2/2025/10}

\begin{document}

\title{HAMLET-FFD: Hierarchical Adaptive Multi-modal Learning Embeddings Transformation for Face Forgery Detection}


\author{Jialei Cui}
\orcid{0009-0000-7923-9329}
\affiliation{%
  \institution{Baidu Inc.}
  \city{Beijing}
  \country{China}
}
\email{cuijialei4009@gmail.com}

\author{Jianwei Du}
\orcid{0000-0001-6358-1904}
\affiliation{%
  \institution{Southeast University}
  \city{Nanjing}
  \country{China}
}
\email{jianwei@seu.edu.cn}

\author{Yanzhe Li}
\orcid{0009-0001-3652-7272}
\affiliation{%
  \institution{Baidu Inc.}
  \city{Beijing}
  \country{China}
}
\email{liyanzhe@baidu.com}

\author{Lei Gao}
\orcid{0009-0000-9475-0280}
\authornote{These authors are co-corresponding authors.}
\affiliation{
  \institution{Baidu Inc.}
  \city{Beijing}
  \country{China}
}
\email{gaolei01@baidu.com}

\author{Hui Jiang}
\orcid{0009-0005-0306-6691}
\authornotemark[1]
\affiliation{
 \institution{Tsinghua University}
 \city{Beijing}
 \country{China}
 }
 \affiliation{
  \institution{Baidu Inc.}
  \city{Beijing}
  \country{China}
}
\email{jianghui01@baidu.com}

\author{Chenfu Bao}
\orcid{0009-0000-6484-1552}
\affiliation{
  \institution{Baidu Inc.}
  \city{Beijing}
  \country{China}
  }
\email{baochenfu@baidu.com}


\begin{abstract}
The rapid evolution of face manipulation techniques poses a critical challenge for face forgery detection: \textit{cross-domain generalization}.
Conventional methods, which rely on simple classification objectives, often fail to learn domain-invariant representations.
We propose HAMLET-FFD, a cognitively inspired Hierarchical Adaptive Multi-modal Learning framework that tackles this challenge via bidirectional cross-modal reasoning.
Building on contrastive vision–language models such as CLIP, HAMLET-FFD introduces a knowledge refinement loop that iteratively assesses authenticity by integrating visual evidence with conceptual cues, emulating expert forensic analysis.
A key innovation is a bidirectional fusion mechanism in which textual authenticity embeddings guide the aggregation of hierarchical visual features, while modulated visual features refine text embeddings to generate image-adaptive prompts.
This closed-loop process progressively aligns visual observations with semantic priors to enhance authenticity assessment.
By design, HAMLET-FFD freezes all pretrained parameters, serving as an external plugin that preserves CLIP’s original capabilities.
Extensive experiments demonstrate its superior generalization to unseen manipulations across multiple benchmarks, and visual analyses reveal a division of labor among embeddings, with distinct representations specializing in fine-grained artifact recognition.
\end{abstract}




\maketitle

\section{Introduction}
Recent advances in artificial intelligence have significantly enhanced facial manipulation techniques, enabling the effortless creation of highly realistic synthetic portraits. 
Meanwhile, these advancements also pose serious threats including identity fraud, misinformation campaigns, and social engineering attacks. 
As forgery techniques rapidly evolve, developing reliable Face Forgery Detection (FFD) methods has become critical for digital media security.

\begin{figure}[tp]
  \centering
  \includegraphics[width=\linewidth]{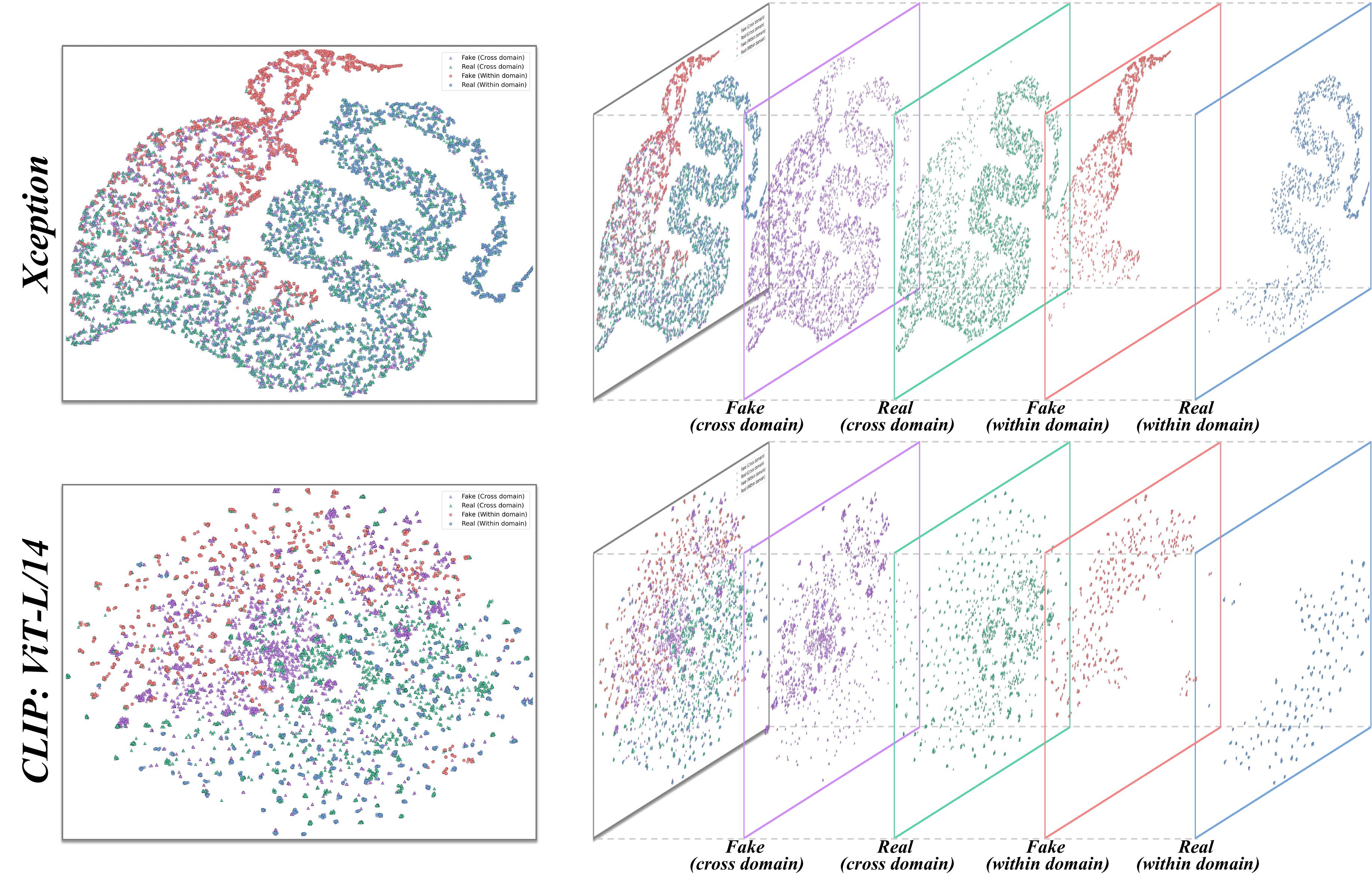}
\caption{Domain-invariant features of CLIP facilitate cross-domain face forgery detection.
(1) Traditional detectors (e.g., Xception) yield entangled features under domain shifts, blurring real/fake boundaries.
(2) In contrast, CLIP provides improved class separation and relatively stable decision boundaries across domains.
This suggests that CLIP implicitly learns general authenticity cues, supporting the design of HAMLET-FFD.
T-SNE visualizations are based on detection features computed as $v * fc.weight + fc.bias$ on \cite{DeepfakeBench_YAN_NEURIPS2023}.}
\label{fig1}
\end{figure}

Traditional face forgery detection approaches predominantly frame the problem as binary classification, typically utilizing CNNs like Xception~\cite{rossler2019faceforensics++} to classify images as real or fake. 
While conceptually straightforward, these methods struggle with a critical challenge: generalizing to unseen manipulation techniques. 
Recent efforts have attempted to improve cross-domain performance by incorporating various complementary signals: frequency-domain features~\cite{qian2020thinking, liu2021spatial, luo2021generalizing}, biological cues~\cite{agarwal2020detecting}, deep forensic representations~\cite{ni2022core, yan2023ucf}, temporal consistency~\cite{afchar2018mesonet, nguyen2019capsule, li2018exposing}, and attention mechanisms~\cite{zhao2021multi}. 
Despite these innovations, significant generalization gaps persist. 
The fundamental limitation stems from the binary classification paradigm itself, which inadvertently encourages models to memorize dataset-specific artifacts rather than learn semantically meaningful representations. 
This shortcut learning behavior results in poor performance when encountering novel forgery techniques outside the training data distribution.

To address these limitations, a new paradigm capturing universal authenticity cues is needed.
Vision-language models (VLMs) offer such a promising shift for face forgery detection. 
Pre-trained on billions of diverse image-text pairs, models like CLIP \cite{radford2021learning} develop rich semantic understanding that transcends domain-specific artifacts.
Figure \ref{fig1} demonstrates this advantage: even a simple linear classifier trained on frozen CLIP features outperforms fully-trained Xception model in separating authentic from manipulated faces, particularly in cross-domain scenarios. 
This reveals that CLIP’s representations inherently encode authenticity cues that generalize better than the domain-specific patterns captured by traditional detectors.

Recent work has started to explore VLMs for face forgery detection, yet much potential is still untapped.
Existing methods primarily fall into three categories: 
(1) those treating CLIP as a fixed feature extractor, like UniFD \cite{ojha2023towards} and Wavelet-CLIP \cite{baru2024harnessing}, which apply linear probing or enhance outputs with frequency analysis; 
(2) prompt-based methods, including CLIPping \cite{khan2024clipping} with learnable context tokens (CoOp~\cite{zhou2022learning}) and VLFFD \cite{sun2023towards} with sentence-level prompts describing facial attributes; 
and (3) parameter-efficient tuning approaches like C2P-CLIP \cite{tan2024c2p}, which applies LoRA to partially fine-tune the image encoder, and RepDFD \cite{lin2024standing}, which learns input-level perturbations. 
Despite these advances, current approaches focus primarily on CLIP’s output representations, overlooking the rich hierarchical features embedded throughout its transformer architecture. 
Furthermore, these methods operate unidirectionally, either adapting a single modality or enforcing strict separation between visual and textual processing.

Building upon these approaches, we observe a critical constraint: the prevailing tendency to frame forgery detection as feature adaptation rather than knowledge extraction.
Our key insight is that CLIP inherently encodes authenticity signals requiring not new learning, but strategic refinement through cross-modal reasoning.
Based on this insight, we present HAMLET-FFD, which advances beyond existing approaches in three fundamental dimensions:
\textit{\textbf{First, HAMLET-FFD harvests multi-level feature representations across CLIP's transformer hierarchy.}} 
Leveraging representations across multiple depths of CLIP’s vision encoder, our framework effectively captures fine-grained artifacts and semantic inconsistencies, facilitating robust detection of diverse manipulation cues.
\textit{\textbf{Second, we implement a bidirectional cross-modal reasoning circuit.}} 
Unlike conventional unidirectional methods, HAMLET-FFD creates reciprocal information pathways where textual concepts guide visual interpretation while visual evidence concurrently refines textual understanding. 
This cognitive feedback loop emulates expert analysis processes where judgment emerges through iterative synthesis of visual cues and conceptual knowledge.
\textit{\textbf{Third, we employ parameter-preserving modular augmentation to maintain CLIP's foundational capabilities.}} 
By freezing CLIP’s pre-trained weights and introducing dedicated forensic reasoning modules, our framework retains semantic robustness while learning manipulation-specific cues, boosting cross-domain performance.

Extensive experiments confirm HAMLET-FFD's superior cross-domain generalization capabilities. 
On seven unseen manipulation techniques from DeepfakeBench~\cite{DeepfakeBench_YAN_NEURIPS2023}, our approach achieves 90.07\% average AUC, surpassing the previous state-of-the-art by a substantial 6.68pp. This advantage extends to emerging forgery techniques, with 92.34\% average AUC on recent benchmarks including diffusion-based manipulations \cite{chen2024diffusionface} and in-the-wild forgeries~\cite{zi2020wilddeepfake, Zhou_2021_CVPR, yan2024df40}. 
Our ablation studies verify each component's contribution, with the bidirectional fusion mechanism providing the most significant performance gains. 
Beyond performance advantages, HAMLET-FFD maintains a modular design that preserves CLIP's original parameters by operating as an external plugin with a fully frozen backbone. 
This architecture enables specialized plugins to be deployed for different scenarios while sharing a common feature extraction backbone, offering practical advantages for real-world deployment.

Our main contributions include:
(1) We propose HAMLET-FFD, a novel framework that leverages CLIP's vision-language knowledge for face forgery detection by treating the task as knowledge refinement rather than feature learning from scratch.
(2) A hierarchical bidirectional fusion mechanism is introduced to model the mutual influence between textual cues and visual features.
(3) Comprehensive experiments demonstrate HAMLET-FFD’s effectiveness on unseen manipulation techniques across multiple benchmarks, with further analysis highlighting the role of bidirectional fusion in enhancing cross-domain generalization.

\section{Related Work} 
\subsection{Face Forgery Detection}
Face forgery detection has evolved significantly in response to increasingly sophisticated manipulation techniques. 
Early approaches relied on generic CNN architectures like Xception \cite{Chollet_2017} and EfficientNet \cite{tan2019efficientnet} for binary classification \cite{rossler2019faceforensics++, afchar2018mesonet}. 
Subsequent research has developed along three main directions: frequency analysis, spatial examination, and generalization-focused methods.
Frequency-domain approaches \cite{durall2020watch, qian2020thinking, liu2021spatial} identify distinctive patterns in synthetic faces' phase spectra and high-frequency components that often reveal manipulation artifacts. 
Complementing these, spatial methods target inconsistencies in challenging facial regions, particularly eyes \cite{Li_Chang_Lyu_2018} and lips \cite{Haliassos_Vougioukas_Petridis_Pantic_2021}.
To address cross-domain generalization, several innovative techniques have emerged. 
Some methods \cite{li2020face, shiohara2022detecting} incorporate synthetic face-swap data during training, while others analyze facial blending artifacts \cite{wang2021representative, chen2022self}. 
Frameworks like UCF \cite{yan2023ucf} extract common forgery features across manipulation techniques, while LSDA \cite{LSDA_YAN_CVPR2024} establishes more generalizable decision boundaries for authenticity verification.
Vision-language models have recently emerged as a promising direction for forgery detection. 
UniFD \cite{ojha2023fakedetect} utilizes frozen CLIP features with nearest neighbor search and linear probing. 
VLFFD \cite{sun2023towards} employs sentence-level prompts describing facial attributes, while Wavelet-CLIP \cite{baru2024harnessing} combines wavelet transformations with CLIP for cross-dataset detection. 
RepDFD \cite{lin2024standing} reprograms CLIP through universal perturbations to visual inputs while preserving internal parameters.
Despite these advances, existing VLM-based methods typically operate unidirectionally and utilize only CLIP's final representations, limiting their ability to perform explicit cross-modal reasoning for robust forgery detection.

\subsection{VLM Adaptation Methods}
Vision-language models (VLMs) align visual and textual features in a shared embedding space and enable flexible zero-shot transfer via natural language prompts~\cite{radford2021learning,jia2021scaling}. However, direct application of generic VLMs such as CLIP~\cite{radford2021learning} to specialized tasks often lags behind task-specific models, while full fine-tuning may degrade robustness to distribution shifts~\cite{Wortsman_2022}. To address this trade-off, parameter-efficient adaptation methods have emerged. CoOp~\cite{zhou2022learning,zhou2022conditional} learns task-specific prompt vectors with frozen encoders, CLIP-Adapter~\cite{Gao_2021} inserts a lightweight residual bottleneck for balanced adaptation, and CLIPping~\cite{khan2024clipping} shows that jointly adapting image and text components benefits universal deepfake detection. These strategies leverage VLMs’ strong semantic priors and cross-domain generalization, making them promising for facial forgery detection.

\section{Methodology}
\label{sec/3_methodology}
\subsection{CLIP as a Backbone}
\label{sec/3_1-clip_as_backbone}
We adopt CLIP~\cite{radford2021learning}, a vision–language model pretrained on 400M image–text pairs, as the foundation of our framework. 
CLIP consists of a vision encoder $\bm{\mathcal{E}}_i(\cdot)$ and a text encoder $\bm{\mathcal{E}}_t(\cdot)$, which map images and text prompts into a shared embedding space where their similarity is measured via cosine distance.
For a given image $\bm{x}$ and a set of class prompts $\{P_y\}$, zero-shot classification is achieved by:
\begin{equation}
p(y|\bm{x}) \propto 
\exp\big( \cos(\bm{\mathcal{E}}_i(\bm{x}), \bm{\mathcal{E}}_t(P_y) )/\tau \big),
\end{equation}
where $\tau$ is a temperature parameter.
In our work, we keep all pretrained parameters frozen and harvest multi-level visual features and text embeddings as inputs to \textbf{HAMLET-FFD}, which performs bidirectional cross-modal reasoning to refine authenticity cues while preserving CLIP’s original semantic knowledge.

\begin{figure*}[t] 
  \centering
  \includegraphics[width=\linewidth]{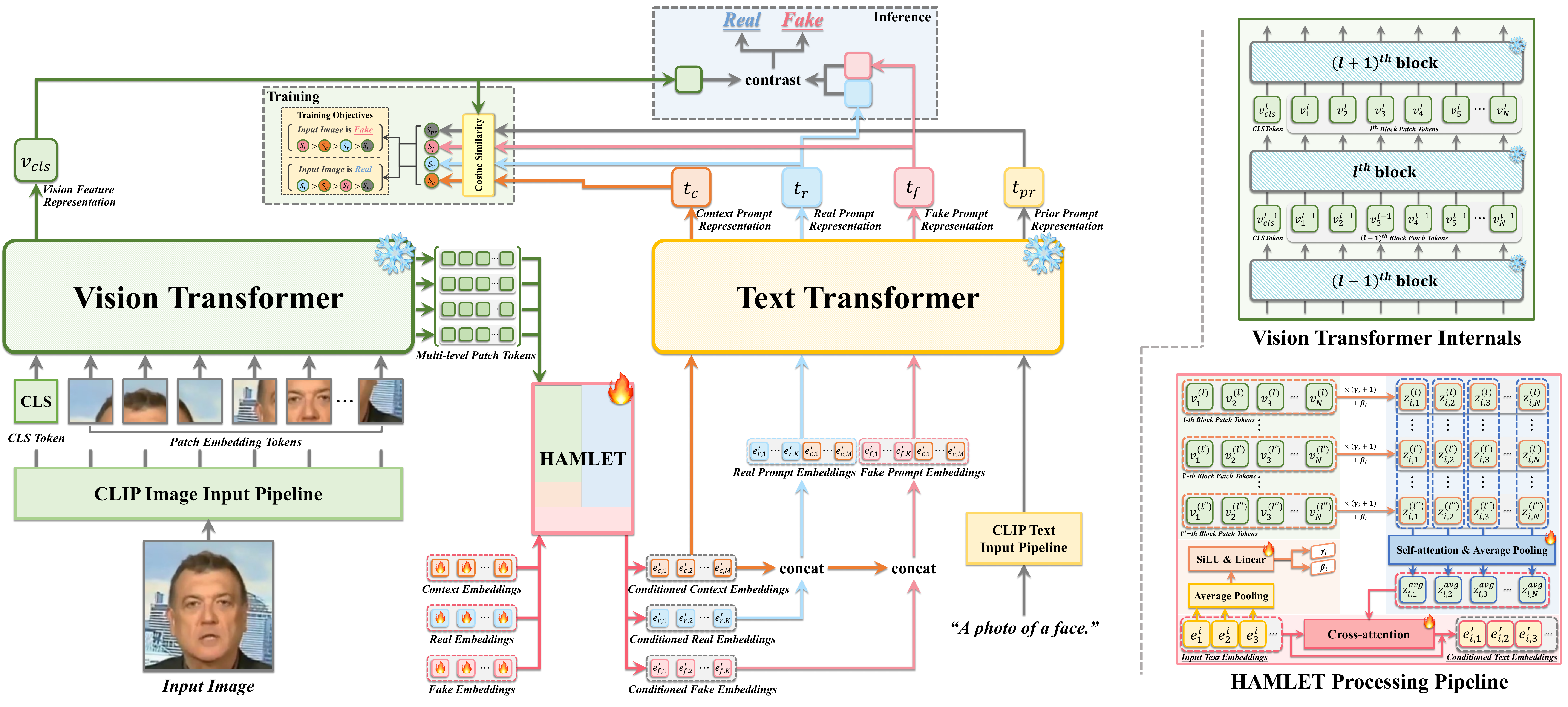}
  \caption{HAMLET-FFD architecture. Left: CLIP's vision transformer (frozen) processes input images and provides multi-level patch tokens from internal blocks. Middle: The HAMLET module (trainable) implements bidirectional modal fusion between visual features and specialized text embeddings. Right: CLIP's text transformer (frozen) processes the conditioned embeddings. The bidirectional fusion occurs in two stages: (1) text embeddings generate scale and shift parameters to modulate multi-level visual features, and (2) aggregated visual features refine text embeddings through cross-attention. During training (top), a hierarchical similarity structure is enforced, while inference uses a simplified comparison between real/fake representations.}
  \label{pipeline}
\end{figure*}

\subsection{Adapting CLIP for Face Forgery Detection}
Having reviewed CLIP's architecture and functionality, we now describe how HAMLET-FFD adapts and extends this foundation specifically for face forgery detection.
As shown in Figure~\ref{pipeline}, HAMLET-FFD builds on CLIP by utilizing its ViT-based visual encoder.
Specifically, the HAMLET module operates as an external plugin that accesses intermediate representations from CLIP's vision transformer and generates vision-conditioned text prompt embeddings for image-adaptive authenticity assessment in the shared CLIP embedding space.
The framework consists of three main components: 
(1) \textit{Hierarchical Visual Feature Access}: Patch embeddings are extracted from multiple blocks of CLIP's vision transformer to utilize its inherent hierarchical representations.
(2) \textit{Specialized Authenticity Embeddings}: Learnable text embeddings are introduced for real faces, fake faces, and shared contextual information that serve as prompts for CLIP's text encoder.
(3) \textit{Bidirectional Modal Fusion}: This component enables interactions between visual features and text embeddings through a two-stage process that conditions text modality with visual information and vice versa.
These components work together to create image-adaptive prompt representations for forgery detection. 
Additionally, HAMLET-FFD maintains CLIP's original parameters, allowing it to be deployed alongside other CLIP-based applications.

\subsubsection{\textbf{Hierarchical Visual Feature Access}}
Face manipulations manifest artifacts at multiple levels—from pixel-level inconsistencies to semantic anomalies. 
For instance, traditional generation-based forgeries (e.g., GAN-based synthesis) often introduce high-frequency texture artifacts detectable in earlier network layers~\cite{durall2020watch, frank2020leveraging, zhang2019detecting}, 
while more sophisticated identity-swapping forgeries tend to cause anatomical inconsistencies or unnatural expressions that are only recognizable in deeper semantic representations~\cite{li2018exposing, yang2019exposing, agarwal2020detecting}.
Relying solely on final-layer representations may miss these multi-level signals.
In HAMLET-FFD, we extract patch embeddings from multiple blocks of CLIP's vision transformer encoder. 
Specifically, given an input image, we pass it through CLIP's visual encoder and collect patch token embeddings $\{v^{(l)}_1, v^{(l)}_2, ..., v^{(l)}_N\}$ from selected blocks $l \in \{l_1, l_2, ..., l_L\}$, where $N$ is the number of image patches and $L$ is the number of selected blocks.
This approach enables the model to simultaneously reason about fine-grained details captured in earlier layers (e.g., texture inconsistencies) and higher-level semantic features in later layers (e.g., anatomical inconsistencies, unnatural expressions). 
The multi-level feature extraction serves as the foundation for our bidirectional modal fusion mechanism.

\subsubsection{\textbf{Specialized Authenticity Embeddings}}
Inspired by CoOp \cite{zhou2022learning}, we introduce three types of learnable textual token embeddings for face forgery detection:
\textit{Real Embeddings} $E_R = \{e^r_1, e^r_2, ..., e^r_{K}\}$ learn to represent characteristics of authentic faces.
\textit{Fake Embeddings} $E_F = \{e^f_1, e^f_2, ..., e^f_{K}\}$ capture distinctive patterns of manipulated faces.
\textit{Context Embeddings} $E_C = \{e^c_1, e^c_2, ..., e^c_{M}\}$ provide task-specific context shared between both categories, where $K$ and $M$ denote the number of tokens for each type.
These embeddings are randomly initialized and optimized during training to develop discriminative representations for authenticity assessment.

\subsubsection{\textbf{Bidirectional Modal Fusion}}
The HAMLET processing pipeline (Figure~\ref{pipeline} bottom right) implements a bidirectional fusion mechanism between visual features and text embeddings. 
Just as humans refine their understanding by interpreting visual cues through language concepts and simultaneously updating those concepts based on what they see, our approach enables hierarchical visual features and textual embeddings to mutually enhance each other through a two-stage process:

\noindent\textit{\textbf{Text-to-Visual Conditioning.}}
First, we condition visual feature interpretation using textual information—analogous to how preconceived concepts guide our visual attention. 
For each type of text embeddings (real, fake, and context), we compute:
\begin{equation}
\mu_i = \text{AvgPool}(E_i) \quad \text{where} \quad i \in \{R, F, C\} \ .
\end{equation}
These averaged embeddings are then transformed through a SiLU activation and linear projection to generate scale and shift parameters:
\begin{equation}
\gamma_i, \beta_i = \text{Linear}(\text{SiLU}(\mu_i)) \ .
\end{equation}
These parameters modulate the multi-level patch embeddings through feature-wise affine transformation:
\begin{equation}
z^{(l)}_{i,j} = v^{(l)}_j \cdot (\gamma_i + 1) + \beta_i \ ,
\end{equation}
where $v^{(l)}_j$ represents the $j$-th patch embedding from the $l$-th transformer block. 
This mechanism injects semantic guidance from each embedding type into visual feature interpretation, helping focus attention on forgery-relevant aspects.

\noindent\textit{\textbf{Self-Attention Integration and Visual-to-Text Conditioning.}}
The text-conditioned patch embeddings from multiple blocks are integrated through self-attention and average pooling:
\begin{equation}
z^{\text{avg}}_{i,j} = \text{AvgPool}(\text{SelfAttention}([z^{(l_1)}_{i,j}, z^{(l_2)}_{i,j}, ..., z^{(l_L)}_{i,j}])) \ .
\end{equation}
In the second stage, these fused visual features condition each type of text embeddings through cross-attention \cite{vaswani2017attention}:
\begin{equation}
E'_i = E_i + \text{CrossAttention}(E_i, Z^{\text{avg}}_i, Z^{\text{avg}}_i) \quad \text{where} \quad i \in \{R, F, C\} \ ,
\end{equation}
with $Z^{\text{avg}}_i = [z^{\text{avg}}_{i,1}, z^{\text{avg}}_{i,2}, ..., z^{\text{avg}}_{i,N}]$ representing the collection of fused visual features.
This bidirectional conditioning creates a closed feedback loop where textual understanding guides visual feature interpretation, and visual evidence refines the textual concepts.
After obtaining the conditioned embeddings $E'_R$, $E'_F$, and $E'_C$, we concatenate them to form the complete prompt embedding sequences:
\begin{equation}
P'_R = [E'_R; E'_C] \quad \text{and} \quad P'_F = [E'_F; E'_C] \ .
\end{equation}
These vision-conditioned prompt embedding sequences are then processed by CLIP's text transformer to produce the final representations used for contrastive learning and classification.

\subsection{Training and Inference}
\subsubsection{\textbf{Training Objective.}}
The bidirectional modal fusion mechanism described above enables HAMLET-FFD to develop image-adaptive representations through cross-modal interaction. 
To effectively leverage this capability, we design a customized training objective that enforces a structured similarity hierarchy, mirroring the progressive refinement of features in the bidirectional fusion process.
As illustrated in Figure~\ref{pipeline}, for each input image, we first extract the visual representation $v_\text{cls}$ using CLIP's visual encoder. 
Meanwhile, our specialized authenticity embeddings are processed through the bidirectional modal fusion module, yielding three conditional prompt representations: $t_r$ (real), $t_f$ (fake), and $t_c$ (context).
Additionally, we employ a fixed prior prompt "A photo of a face." that yields a prior representation $t_{pr}$, serving as a regularization signal throughout training.
For a given image, we compute temperature-scaled cosine similarities with each prompt representation:
\begin{equation}
\begin{aligned}
s_r = \cos(v_\text{cls}, t_r) / \tau \ ;\ \ \  & \ \ \ s_f = \cos(v_\text{cls}, t_f) / \tau \ ; \\
s_c = \cos(v_\text{cls}, t_c) / \tau \ ;\ \ \  & \ \ \ s_{pr} = \cos(v_\text{cls}, t_{pr}) / \tau \ ,
\end{aligned}
\end{equation}
where $\tau$ is CLIP's learned temperature parameter.
We organize these similarity measures differently depending on the authenticity of the image. 
For a real image, the similarities are arranged as $S = [s_r, s_c, s_f, s_{pr}]$, whereas for a fake image, they follow the order $S = [s_f, s_c, s_r, s_{pr}]$.
This arrangement places the expected highest similarity first, followed by decreasing levels of relevance. 
We then compute a progressive cross-entropy loss:
\begin{equation}
\mathcal{L} = \sum_{i=0}^{2} (0.5)^i \cdot \text{CrossEntropy}(S[i:], 0) \ ,
\end{equation}
where $S[i:]$ represents the subsequence of similarities starting from position $i$, and $0$ indicates that the first position in each subsequence should have the highest similarity. 
The weighting factor $(0.5)^i$ assigns decreasing importance to the loss computed on later subsequences.
This formulation enforces a clear hierarchy of similarities ($s_r > s_c > s_f > s_{pr}$ for real images and $s_f > s_c > s_r > s_{pr}$ for fake images), helping the model learn more nuanced representations for face authenticity assessment.

\subsubsection{\textbf{Inference Procedure.}}
During inference, the procedure is simplified to focus on the essential binary decision. 
Given an input image, we extract its visual representation $v_\text{cls}$ and compute its similarities only with the conditioned real and fake prompt representations:
\begin{equation}
s_r = \cos(v_\text{cls}, t_r)  \quad \text{and} \quad s_f = \cos(v_\text{cls}, t_f)  \ .
\end{equation}
The classification decision is made by directly comparing these two similarity scores:
\begin{equation}
\text{Prediction} = 
\begin{cases}
\text{Real}, & \text{if } s_r > s_f \ ; \\
\text{Fake}, & \text{otherwise} \ .
\end{cases}
\end{equation}

\section{Experiments}
\label{sec/4_experiments}
\subsection{Experimental Setup}
\label{sec/4_1-experimental_settings}
\noindent\textit{\textbf{Datasets.}}
We conduct our primary experiments on multiple popular datasets curated by DeepfakeBench \cite{DeepfakeBench_YAN_NEURIPS2023}.
Following its evaluation protocol, we adopt FaceForensics++ (FF++) \cite{rossler2019faceforensics++} as the training set, which contains 1,000 pristine videos manipulated by four representative techniques: Deepfakes (FF-DF) \cite{DeepFakes}, Face2Face (FF-F2F) \cite{thies2016face2face}, FaceSwap (FF-FS) \cite{FaceSwap}, and NeuralTextures (FF-NT) \cite{thies2019deferred}.
We follow the standard c23 setting \cite{chen2022self,Chen_Zhang_Song_Wang_Liu,li2020face}, representing a moderate compression level.
For evaluation, we consider a broad set of widely-used benchmarks: CelebDF (v1 and v2) \cite{li2020celeb}, DeepFakeDetection (DFD) \cite{DFD}, the DeepFake Detection Challenge (DFDC) \cite{dolhansky2020deepfake} and its preview version (DFDC-P) \cite{guarnera2022face}, UADFV \cite{li2018ictu}, and FaceShifter (Fsh) \cite{li2020advancing}.
To further assess generalizability under recent and challenging manipulations, we extend our evaluation to four additional benchmarks: WildDeepfake (WDF) \cite{zi2020wilddeepfake}, FFIW \cite{Zhou_2021_CVPR}, DiffusionFace \cite{chen2024diffusionface}, and the recently introduced DF40 \cite{yan2024df40}.
For DiffusionFace and DF40, which include multiple manipulation types, we follow \cite{cheng2024can} and select representative subsets: DiffSwap from DiffusionFace, and UniFace, E4S, BlendFace, and MobileSwap from DF40.
These extended benchmarks capture emerging paradigms such as diffusion-based synthesis and mobile-device-generated forgeries.

\noindent\textit{\textbf{Evaluation metrics.}}
Following the official evaluation protocol \cite{DeepfakeBench_YAN_NEURIPS2023}, we use frame-level Area Under Curve (AUC) as the primary metric for comparing against state-of-the-art detectors. 
Additional evaluation results, including accuracy (ACC), average precision (AP), equal error rate (EER), and video-level AUC, are provided in the supplementary material for a more comprehensive analysis.

\noindent\textit{\textbf{Baselines.}}
We compare HAMLET-FFD with state-of-the-art methods included in DeepfakeBench~\cite{DeepfakeBench_YAN_NEURIPS2023}, 
covering three major technical directions: CNN-based architectures~\cite{cao2022end, ni2022core, LSDA_YAN_CVPR2024}, 
frequency-domain analysis~\cite{qian2020thinking, liu2021spatial, luo2021generalizing}, 
and feature disentanglement~\cite{yan2023ucf, cheng2024can}. 
We additionally reproduce recent CLIP-based approaches~\cite{ojha2023fakedetect, sun2023towards, baru2024harnessing, khan2024clipping, tan2024c2p, lin2024standing} using official weights or verified implementations, 
enabling a direct comparison of our bidirectional fusion strategy against methods leveraging similar foundation models but with different adaptation schemes.

\noindent\textit{\textbf{Implementation details.}}
Following prior works~\cite{ojha2023fakedetect, sun2023towards, baru2024harnessing, khan2024clipping, tan2024c2p, lin2024standing}, 
we adopt CLIP:ViT-L/14 as the backbone. 
The authenticity embeddings are initialized with $(2,2,16)$ tokens for the Real, Fake, and Context categories, respectively. 
Our bidirectional modal fusion employs a 32-head multi-head attention module, selected as the optimal trade-off between computational cost and detection accuracy. 
For hierarchical visual access, patch embeddings are extracted from every 4th block of CLIP's vision transformer (blocks 4, 8, 12, 16, 20, and 24), yielding six feature levels. 
All experiments are conducted on the DeepfakeBench framework~\cite{DeepfakeBench_YAN_NEURIPS2023} using its default settings, except that face images are resized from $256{\times}256$ to $224{\times}224$ to match CLIP's input specification. 
Training is performed on two NVIDIA A100 GPUs for 20 epochs with the AdamW optimizer (learning rate $1\mathrm{e}{-3}$), and the temperature parameter $\tau$ is directly inherited from CLIP's pretrained model buffer.

\begin{table*}[!ht]
    \caption{Cross-domain generalization performance comparison on DeepfakeBench benchmark (AUC\%). All models are trained on FF++ (c23) and evaluated on other datasets. The table presents both within-domain results (same distribution) and cross-domain results (unseen manipulation techniques). Traditional approaches (top) are compared with CLIP-based methods (highlighted in \colorbox{clip_red}{red}). Models marked with $\dagger$ use official pre-trained weights or verified reproductions. Best and second-best results are highlighted in \textbf{bold} and \underline{underlined} respectively.}
        \label{tab:deepfakebench-auc}
        \centering
        \resizebox{\textwidth}{!}{
        \begin{tabular}{l | c | c c c c c c | c c c c c c c c}
        \toprule
            \multirow{2}*{\textbf{Methods}}  & \multirow{2}{*}{\textbf{Venues}} & \multicolumn{6}{c|}{\textbf{Within Domain Evaluation}} &  \multicolumn{8}{c}{\textbf{Cross Domain Evaluation}} \\
            \cmidrule(lr){3-8} \cmidrule(lr){9-16}
            ~                                                                   & ~   & \makecell[c]{FF++} & \makecell[c]{FF-DF} & \makecell[c]{FF-F2F} & \makecell[c]{FF-FS}  & \makecell[c]{FF-NT}  & \makecell[c]{Avg.}  & \makecell[c]{CDFv1}& \makecell[c]{CDFv2}& \makecell[c]{DFD}& \makecell[c]{DFDC}& \makecell[c]{DFDCP}& \makecell[c]{Fsh} & \makecell[c]{UADFV} & \makecell[c]{Avg.} \\
            \midrule
            SPSL \cite{liu2021spatial}                                          & CVPR'20       & 96.10             & 97.81             & 97.54             & 98.29             & 92.99             & 94.08             & 81.50             & 76.50             & 81.22             & 70.40             & 74.08             & 64.37             & 94.24             & 78.75             \\
            F3Net \cite{qian2020thinking}                                       & ECCV'20       & 96.35             & 97.93             & 97.96             & 98.44             & 93.54             & 94.49             & 77.69             & 73.52             & 79.75             & 70.21             & 73.54             & 59.14             & 93.47             & 76.45             \\
            SRM \cite{luo2021generalizing}                                      & CVPR'21       & 95.76             & 97.33             & 96.96             & 97.40             & 92.95             & 93.59             & 79.26             & 75.52             & 81.20             & 69.95             & 74.08             & 60.14             & 94.27             & 77.60             \\
            Recce \cite{cao2022end}                                             & CVPR'22       & 96.21             & 97.97             & 97.79             & 97.85             & 93.57             & 94.22             & 76.77             & 73.19             & 81.19             & 71.33             & 74.19             & 60.95             & 94.46             & 76.49             \\
            CORE \cite{ni2022core}                                              & CVPRW'22      & 96.38             & 97.87             & 98.03             & 98.23             & 93.39             & 94.31             & 77.98             & 74.28             & 80.18             & 70.49             & 73.41             & 60.32             & 94.12             & 76.94             \\
            UCF \cite{yan2023ucf}                                               & ICCV'23       & 97.05             & 98.83             & 98.40             & \textbf{98.96}    & 94.41             & 95.27             & 77.93             & 75.27             & 80.74             & 71.91             & 75.94             & 64.62             & 95.28             & 78.01             \\
            LSDA \cite{LSDA_YAN_CVPR2024}                                       & CVPR'24       & 93.89             & 93.32             & 94.46             & 96.26             & 93.14             & 94.21             & 86.70             & 83.01             & 88.02             & 73.60             & 81.52             & 67.63             & 83.19             & 82.67             \\
            ProDet \cite{cheng2024can}                                          & NeurIPS'24    & 95.19             & 96.05             & 95.80             & 95.34             & 93.81             & 95.24             & \textbf{90.94}    & \underline{84.48} & 85.81             & 72.40             & 81.16             & 64.53             & 91.17             & 81.50             \\
            {FA-ViT$\dagger$} \cite{favit}                                                 & TCSVT'24      & 98.13             & \textbf{99.48}    & 97.95             & 98.23             & 96.86             & 98.13             & 78.04             & 83.29             & \underline{88.50}             & 76.03             & 81.16             & 65.19             & 88.43             & 80.09             \\
            \rowcolor{clip_red}
            {UniFD$\dagger$} \cite{ojha2023fakedetect}         & CVPR'23       & 85.39             & 96.77             & 83.96             & 93.97             & 67.56             & 85.53             & 74.21             & 72.84             & 84.33             & 73.98             & 76.11             & 76.04             & 95.26             & 78.97             \\
            \rowcolor{clip_red}
            {VLFFD$\dagger$} \cite{sun2023towards}             & arXiv'23      & \textbf{98.53}    & \underline{99.37} & \textbf{99.19}    & \underline{98.70} & \underline{97.02} & \textbf{98.56}    & 71.69             & 75.14             & 79.82             & 73.97             & 76.21             & 70.24             & 87.49             & 76.37             \\
            \rowcolor{clip_red}
            {CLIPping$\dagger$} \cite{khan2024clipping}        & ICMR'24       & 83.86             & 96.86             & 80.89             & 94.82             & 63.39             & 83.96             & 79.02             & 75.69             & 85.91             & 75.59             & 74.79             & \underline{77.37} & 96.86             & 80.75             \\
            \rowcolor{clip_red}
            {Wavelet-CLIP$\dagger$} \cite{baru2024harnessing}  & WACV'25       & 88.48             & 97.39             & 88.63             & 94.76             & 74.28             & 88.71             & 75.61             & 75.93             & 85.98             & 75.04             & 77.81             & 73.20             & 93.79             & 79.62             \\
            \rowcolor{clip_red}
            {C2P-CLIP$\dagger$} \cite{tan2024c2p}              & AAAI'25       & \underline{98.50} & 99.30             & \underline{98.99} & 98.54             & \textbf{97.18}    & \underline{98.50} & 74.63             & 80.20             & 88.29             & \underline{78.71} & 82.26             & 71.23             & \underline{97.46} & 81.83             \\
            \rowcolor{clip_red}
            {RepDFD$\dagger$} \cite{lin2024standing}           & AAAI'25       & 96.72             & 97.48             & 96.04             & 96.86             & 90.16             & 95.45             & 84.07             & 82.29             & 84.84             & 76.95             & \underline{82.56} & 77.11             & 95.98             & \underline{83.39} \\
            \midrule
            \rowcolor{our_blue} 
            \textbf{Ours}                                   & ---            & 97.68             & 99.30             & 98.12             & 98.60             & 94.93             & 97.73             & \underline{90.46} & \textbf{89.46}    & \textbf{94.35}    & \textbf{85.39}    & \textbf{89.60}    & \textbf{82.29}    & \textbf{98.95}    & \textbf{90.07}    \\
            \bottomrule
        \end{tabular}
        }
        \end{table*}

\subsection{Cross-Domain Generalization}
\label{sec/4_2-cross_domain_generalization}
\subsubsection{\textbf{Performance on DeepfakeBench}}
Table \ref{tab:deepfakebench-auc} presents a comprehensive performance comparison on the DeepfakeBench. 
While our method does not achieve the highest within-domain score (97.73\% AUC), this result reflects a deliberate design choice to freeze the pre-trained backbone and prioritize generalization in cross-domain settings—the primary focus of our work.
HAMLET-FFD achieves 90.07\% average AUC across seven cross-domain benchmarks, surpassing the best-performing baseline (RepDFD) by a substantial margin of 6.68pp. 
The performance gap is particularly pronounced on challenging datasets like DFDC (85.39\% vs. 78.71\%) and DFDCP (89.60\% vs. 82.56\%), which contain a wide variety of manipulation techniques created under diverse conditions.
Notably, while some CLIP-based methods like VLFFD and C2P-CLIP excel in within-domain evaluation (achieving 98.56\% and 98.50\% average AUC respectively), their performance degrades significantly in cross-domain scenarios. 
This notable performance gap between within-domain and cross-domain scenarios reveals that these methods excel at learning dataset-specific patterns but struggle to capture universal forgery characteristics.
In contrast, HAMLET-FFD, with its bidirectional fusion mechanism, maintains more consistent performance across both settings, indicating its superior ability to extract domain-invariant representations.

\subsubsection{\textbf{Generalization to Recent Forgery Techniques}}
To further challenge our method, we evaluate its performance on more recent and diverse manipulation techniques not included in the original DeepfakeBench framework. 
Table \ref{tab:cross-benchmark-auc} shows that HAMLET-FFD consistently outperforms all baselines across these extended benchmarks, achieving 92.34\% average AUC compared to 86.84\% for the next best method.
The performance improvement is particularly significant on WildDeepfake (88.44\% vs. 82.80\%) and FFIW (86.60\% vs. 82.48\%), which contain in-the-wild forgeries captured under uncontrolled conditions. 
Notably, our method also excels on DiffSwap (99.03\%), demonstrating effective generalization to modern diffusion-based manipulation techniques that were developed years after CLIP's pre-training.
On the challenging DF40 benchmark, which encompasses diverse real-world manipulations like BlendFace and MobileSwap, HAMLET-FFD maintains its substantial lead with consistent performance improvements across all subsets. 
The ability to generalize across such diverse manipulation techniques—from traditional computer graphics approaches to cutting-edge generative models—validates our bidirectional fusion approach for capturing universal authenticity cues rather than technique-specific artifacts.

\begin{figure*}[tp]
  \centering
  \includegraphics[width=\linewidth]{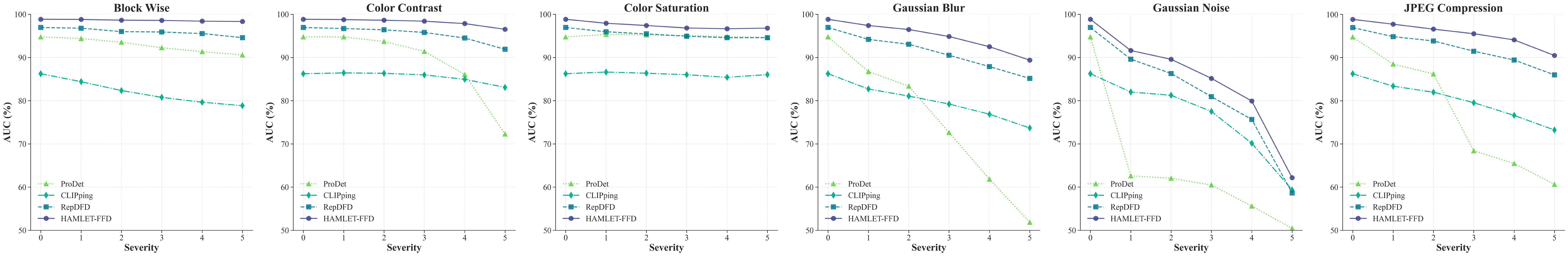}
  \caption{Robustness comparison under various perturbations. Each plot shows video-level AUC (\%) versus perturbation severity (0-5) for different methods. All methods are trained on the FF++ c23 dataset.}
  \label{fig:robustness}
\end{figure*}

\subsection{Ablation Studies}
\label{sec/4_3-ablation_studies}
\noindent\textit{\textbf{Robustness Analysis.}}
Following \cite{jiang2020deeperforensics}, we evaluate HAMLET-FFD's robustness to various perturbations compared to state-of-the-art methods (ProDet, CLIPping, and RepDFD) in video-level AUC. 
We test performance against six types of perturbations across five severity levels. 
Figure~\ref{fig:robustness} shows that HAMLET-FFD maintains consistently higher AUC across all perturbation types and severity levels. 
The performance gap becomes particularly pronounced at higher severity levels, where other methods show significant degradation. 

\begin{table}[tp]
    \caption{Performance on emerging forgery techniques (AUC\%). Models trained only on FF++ (c23) are evaluated on recent benchmarks representing state-of-the-art manipulation methods: WildDeepfake \cite{zi2020wilddeepfake}, FFIW \cite{Zhou_2021_CVPR}, DiffSwap \cite{chen2024diffusionface}, and DF40 \cite{yan2024df40} subsets. Best and second-best results are highlighted in \textbf{bold} and \underline{underlined} respectively.}
        \label{tab:cross-benchmark-auc}
        \centering
        \resizebox{\columnwidth}{!}{
        \begin{tabular}{l | c | c | c | c c c c | c}
        \toprule
            \multirow{2}*{\textbf{Methods}}   & \multirow{2}{*}{\textbf{WDF}} & \multirow{2}{*}{\textbf{FFIW}} & \multirow{2}{*}{\textbf{DiffSwap}} & \multicolumn{4}{c|}{\textbf{DF40}}                                                                                                            & \multirow{2}{*}{\textbf{Avg.}} \\
            \cmidrule(lr){5-8} 
            ~                                  & ~                            & ~                              & ~                                  & \makecell[c]{\footnotesize{UniFace}} & \makecell[c]{\footnotesize{E4S}} & \makecell[c]{\footnotesize{BlendFace}} & \makecell[c]{\footnotesize{MobileSwap}} & ~                  \\
            \midrule
            UCF                                & 69.89                        & 63.32                          & 80.33                              & 83.55                                & 63.55                            & 79.38                                  & 84.42                                   & 74.92              \\
            LSDA                               & 77.61                        & 68.80                          & 83.69                              & 74.67                                & 80.09                            & 71.34                                  & 88.38                                   & 77.80              \\
            ProDet                             & 77.18                        & 70.40                          & 84.59                              & 84.41                                & 71.03                            & \underline{86.19}                      & 92.85                                   & 80.95              \\
            FA-ViT                             & 82.53                        & 80.74                          & 94.58                              & \underline{89.19}                    & 73.69                            & 83.26                                  & 79.30                                   & 83.33              \\
            \rowcolor{clip_red}CLIPping        & 70.93                        & \underline{82.48}              & 97.26                              & 80.85                                & \underline{91.96}                & 75.73                                  & 86.36                                   & 83.65              \\
            \rowcolor{clip_red}Wavelet-CLIP    & 74.23                        & 82.29                          & \underline{97.82}                  & 85.05                                & 87.50                            & 74.72                                  & 75.58                                   & 82.46              \\
            \rowcolor{clip_red}C2P-CLIP        & \underline{82.80}            & 72.13                          & 97.22                              & 86.23                                & 89.34                            & 81.48                                  & 90.56                                   & 85.68              \\
            \rowcolor{clip_red}RepDFD          & 81.84                        & 76.95                          & 97.25                              & 86.17                                & 91.10                            & 81.27                                  & \underline{93.33}                       & \underline{86.84}  \\
            \midrule
            \rowcolor{our_blue} \textbf{Ours}  & \textbf{88.44}               & \textbf{86.60}                 & \textbf{99.03}                     & \textbf{91.99}                       & \textbf{95.71}                   & \textbf{89.43}                         & \textbf{95.21}                          & \textbf{92.34}     \\
            \bottomrule
        \end{tabular}
        }
    \end{table}

\noindent\textit{\textbf{Bidirectional Fusion Mechanism.}}
Table~\ref{tab:ablation_fusion} demonstrates how each component of our bidirectional fusion mechanism contributes to cross-domain generalization. 
Replacing the cross-attention in the V$\rightarrow$T pathway (visual-to-text conditioning) with direct pooling and addition leads to the most severe performance drop ($-$3.90pp cross-domain AUC), highlighting the importance of visual guidance in refining textual embeddings.
When reduced to prompt tuning only (removing all fusion components), performance drops dramatically ($-$9.88pp), validating that our bidirectional approach is fundamental to HAMLET-FFD's generalization capability.

\begin{figure}[tp]
  \centering
  \includegraphics[width=\linewidth]{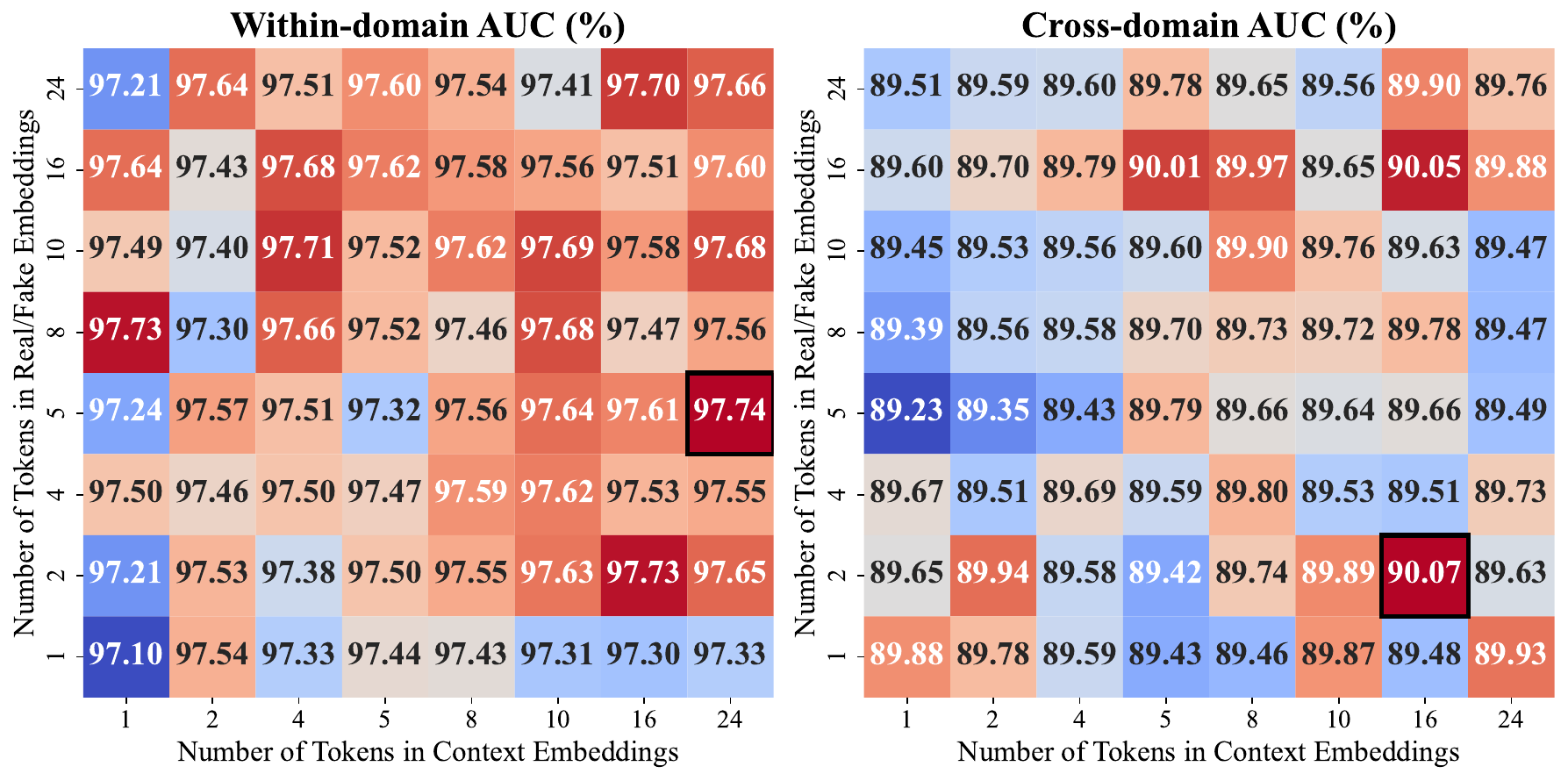}
  \vspace{-0.275in}
  \caption{Grid Search across Token Counts. Heatmaps showing AUC performance with varying numbers of Context tokens (x-axis) and Real/Fake tokens (y-axis). The optimal configuration (16, 2) achieves 90.07\% cross-domain AUC (right).}
  \label{fig:token_heatmap}
\end{figure}

\noindent\textit{\textbf{Text Embedding Configuration.}}
We analyze two key aspects of text embeddings: sequence arrangements and composition strategies. 
As shown in Table~\ref{tab:ablation_tokens}, different embedding arrangements have relatively minor impact on performance (varying within 0.35pp), with the configuration ([R,F] + [C]) performing slightly better. 
In contrast, embedding composition significantly affects generalization.
Removing \textit{Context Embeddings} entirely causes substantial degradation ($-$3.20pp), confirming their crucial role in capturing domain-invariant features. 
Figure~\ref{fig:token_heatmap} visualizes our grid search over token quantities for each embedding category, showing that using 2 Real/Fake tokens and 16 Context tokens achieves an optimal trade-off between efficiency and performance.

\noindent\textit{\textbf{Hierarchical Feature Extraction.}}
Table~\ref{tab:ablation_layers} demonstrates the impact of multi-level feature extraction from CLIP's vision transformer. 
Using only the final block significantly reduces performance ($-$4.17pp cross-domain AUC compared to our default setting), confirming that different levels capture complementary forgery cues. 
Our configuration (every 4th block) achieves an optimal balance between detection performance (90.07\%) and computational efficiency.

\noindent\textit{\textbf{Training Objective.}}
Table~\ref{tab:ablation_loss} evaluates different variants of our training objective. 
The structured similarity hierarchy approach with progressive weighting achieves 90.07\% cross-domain AUC, substantially outperforming standard cross-entropy classification ($+$2.09pp). 
The prior prompt regularization component contributes an additional improvement of 0.69pp, suggesting that our complete objective design helps the model better distinguish authentic from manipulated faces across different domains.
\begin{table}[t]\footnotesize
\caption{Ablation study on bidirectional fusion components. Each component contributes to cross-domain generalization, with the visual$\rightarrow$text pathway being most critical.}
\label{tab:ablation_fusion}
\centering
\resizebox{\columnwidth}{!}{
\begin{tabular}{l|ccc|cc}
\toprule
\multirow{2}{*}{Model Variant} & \multicolumn{3}{c|}{Components} & \multicolumn{2}{c}{AUC (\%)} \\
\cmidrule(lr){2-4} \cmidrule(lr){5-6}
 & T$\rightarrow$V & SA & V$\rightarrow$T & Within & Cross \\
\midrule
\rowcolor{gray!15}
\textit{Full model} & \ding{51} & \ding{51} & \ding{51} & \textbf{97.73} & \textbf{90.07} \\
\textit{w/o T$\rightarrow$V} & \ding{55} & \ding{51} & \ding{51} & 97.44 & 89.05 \\
\textit{w/o SA} & \ding{51} & \ding{55} & \ding{51} & 96.99 & 88.56 \\
\textit{w/o V$\rightarrow$T} & \ding{51} & \ding{51} & \ding{55} & 96.87 & 86.17 \\
\textit{w/o T$\rightarrow$V \& SA} & \ding{55} & \ding{55} & \ding{51} & 96.83 & 87.23 \\
\textit{Prompt tuning only} & \ding{55} & \ding{55} & \ding{55} & 83.01 & 80.19 \\
\bottomrule
\end{tabular}
}
\vspace{0.005em}

\scriptsize
\textit{T$\rightarrow$V}: Text-to-visual conditioning;  \textit{SA}: Self-attention integration;  \textit{V$\rightarrow$T}: Visual-to-text conditioning; \\
``\textit{w/o V$\rightarrow$T}": Replacing cross-attention with direct pooling and addition. \ \ \ \ \ \ \ \ \ \ \ \ \ \ \ \ \ \ \ \ \ \ \ \ \ \ \ \ \ \ \ \ \ \ \ \ \ \ \ \ \ \ \ \ \ 
\end{table}
\begin{table}[tp]\footnotesize
\centering
\caption{Ablation study on text Embedding configurations. Comparison of embedding arrangements and composition strategies. Our default configuration ([R,F]+[C] with all learnable tokens) achieves optimal cross-domain generalization.}
\label{tab:ablation_tokens}
\resizebox{\columnwidth}{!}{
\begin{tabular}{l|
>{\columncolor{gray!15}}c
c
c|
>{\columncolor{gray!15}}c
c
c}
\toprule
\multirow{2}{*}{\makecell[c]{AUC \\ (\%)}} & \multicolumn{3}{c|}{\small{Sequence Arrangements}} & \multicolumn{3}{c}{\small{Composition Strategies}} \\
\cmidrule(lr){2-4} \cmidrule(lr){5-7}
& \scriptsize{[R,\ F]+[C]} & \scriptsize{[C]+[R,\ F]} & \scriptsize{Split} & \scriptsize{All learnable} & \scriptsize{No context} & \scriptsize{Fixed [R,\ F]} \\
\midrule
{Within} & \small{97.73} & \small{\textbf{97.94}} & \small{97.65} & \small{97.73} & \small{96.38} & \small{97.12} \\
{Cross} & \small{\textbf{90.07}} & \small{89.91} & \small{89.72} & \small{\textbf{90.07}} & \small{86.87} & \small{88.29} \\
\bottomrule
\end{tabular}
}
\vspace{0.005em}

\scriptsize
{[R,\ F]}: Real/Fake embeddings; \ \ \ \ \ \ \ \ {[C]}: Context embeddings; \ \ \ \ \ \ \ \ Split: [C]+[R,\ F]\ +\ [C] \ \ \ \ \ \ \ \  \ \ \ \ \ \ \ \  \ \ \ \ \ \ \ \  \\
``Fixed {[R,\ F]}": Using CLIP's text embeddings for ``real” and ``fake”, instead of learnable embeddings.
\end{table}
\begin{table}[tp]\footnotesize
\centering
\caption{Ablation study on hierarchical feature extraction. Performance drops significantly with fewer feature levels, highlighting the importance of multi-level representations.}
\label{tab:ablation_layers}
\resizebox{\columnwidth}{!}{
\begin{tabular}{l|ccc|c|cccc}
\toprule
\multirow{3}{*}{\shortstack[c]{\small AUC \\ \small (\%)}} 
  & \multicolumn{8}{c}{\small Feature Source (Number of Levels)} \\
\cmidrule(lr){2-9}
& \scriptsize{All} & \scriptsize{Every 2nd} & \scriptsize{Every 3rd} & \cellcolor{gray!15}\scriptsize{Every 4th} & \scriptsize{Every 6th} & \scriptsize{Every 8th} & \scriptsize{Every 12th} & \scriptsize{Final} \\
& (24) & (12) & (8) & \cellcolor{gray!15}(6) & (4) & (3) & (2) & (1) \\
\midrule
{Within} & \small{\textbf{98.32}} & \small{97.96} & \small{97.95} & \cellcolor{gray!15}\small{97.73} & \small{97.36} & \small{96.59} & \small{95.55} & \small{95.18} \\
{Cross} & \small{\textbf{90.42}} & \small{90.21} & \small{89.60} & \cellcolor{gray!15}\small{90.07} & \small{89.64} & \small{88.26} & \small{86.98} & \small{85.90} \\
\midrule
MFLOPs & \small{1334.65} & \small{1287.47} & \small{1271.74} & \cellcolor{gray!15}\small{1263.87}  & \small{1256.01} & \small{1252.08} & \small{1248.14} & \small{1244.21} \\
\bottomrule
\end{tabular}
}
\end{table}
\begin{table}[tp]\footnotesize
\caption{Ablation study on training objectives. Our complete objective with structured similarity hierarchy and prior prompt regularization outperforms standard classification approaches, especially for cross-domain generalization.}
\label{tab:ablation_loss}
\centering
\renewcommand{\arraystretch}{0.7}
\resizebox{\columnwidth}{!}{
\begin{tabular}{l|cc}
\specialrule{0.08em}{0pt}{1pt}  
\multirow{2}{*}{\tiny{Training Objective}} & \multicolumn{2}{c}{\tiny{AUC (\%)}} \\
\cmidrule(lr){2-3}
& \tiny{Within} & \tiny{Cross} \\
\midrule
\rowcolor{gray!15}
\tiny{\textit{Structured similarity hierarchy} (complete)} & \tiny{\textbf{97.73}} & \tiny{\textbf{90.07}} \\
\tiny{\textit{w/o Prior prompt regularization}} & \tiny{97.53} & \tiny{89.38} \\
\tiny{\textit{Standard cross-entropy}} & \tiny{97.25} & \tiny{87.98} \\
\specialrule{0.08em}{1pt}{0pt}  
\end{tabular}
}
\end{table}
\begin{figure*}[t] 
  \centering
  \includegraphics[width=0.97\linewidth]{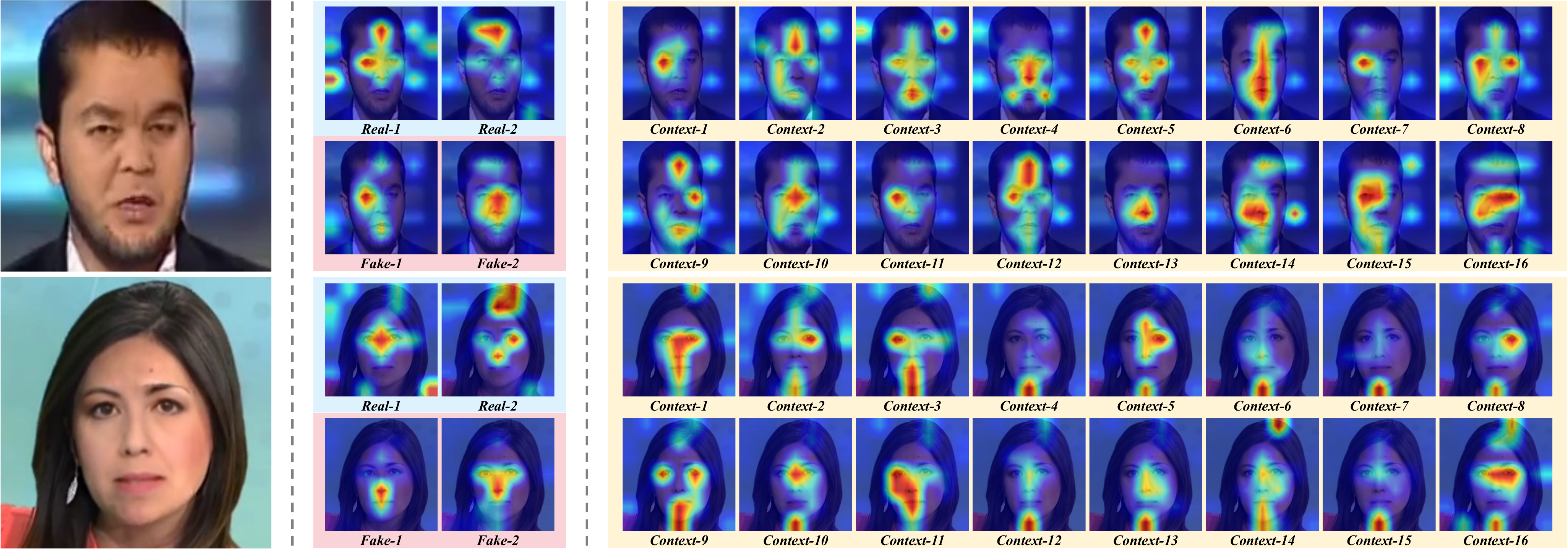}
  \vspace{-0.1in}
  \caption{Attention visualization across embedding categories on manipulated faces. Real embeddings (top-left) attend to global facial symmetry, while Fake embeddings (bottom-left) focus on localized artifacts (e.g., eyes, mouth corners). Context embeddings (right) exhibit distributed specialization, with tokens attending to complementary facial regions. This forms a dynamic ensemble of detectors that adaptively assess authenticity. Data from FF++ (containing AI-generated forgeries).}
  \label{attention}
\end{figure*}
\subsection{Interpretability and Mechanistic Insights}
Beyond strong performance, understanding how HAMLET-FFD works offers deeper insights. Our visualizations highlight the role of cross-modal reasoning in narrowing the generalization gap.
\begin{figure}[t]
  \centering
  \includegraphics[width=\linewidth]{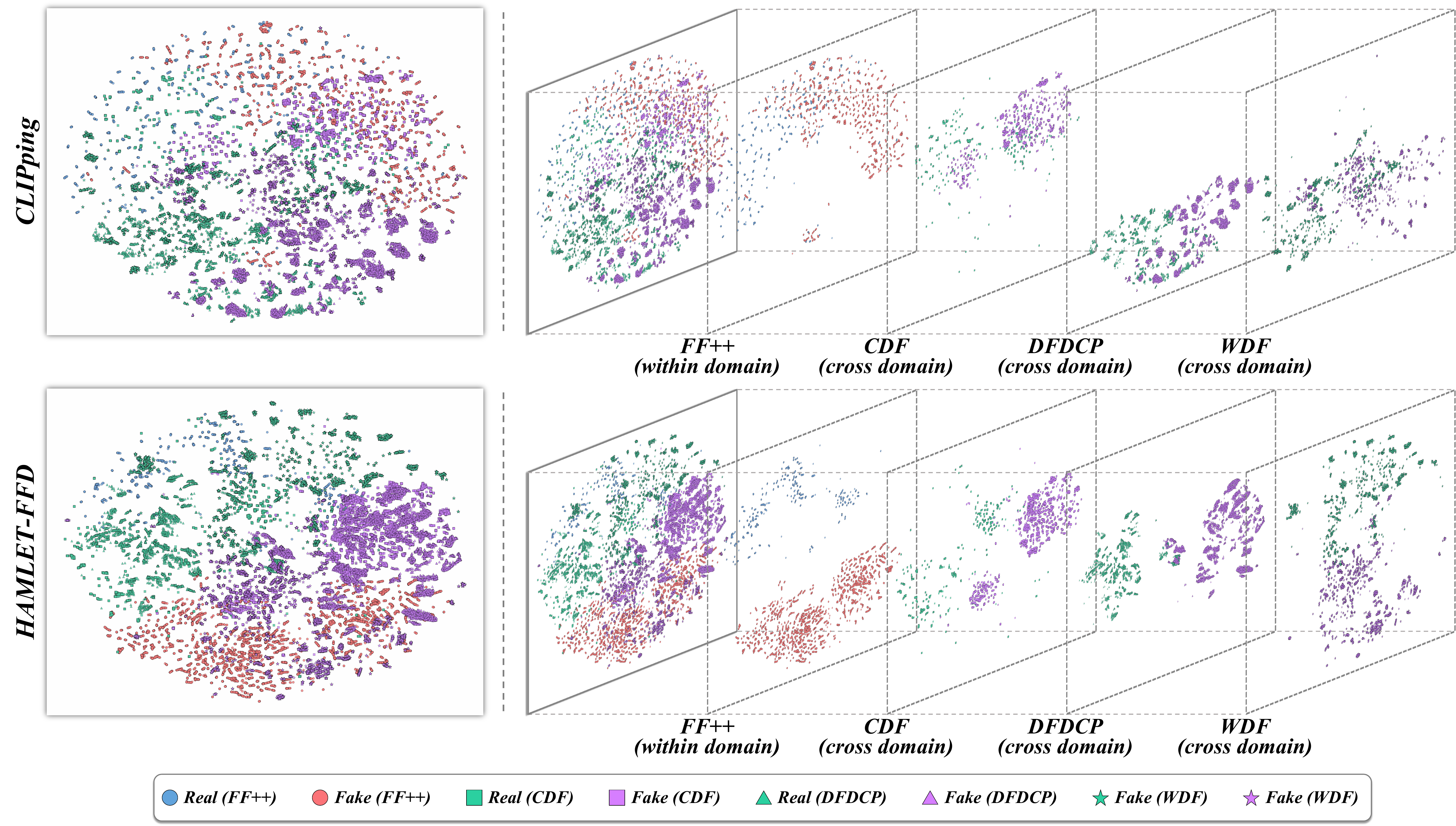}
  \vspace{-0.2in}
  \caption{t-SNE visualization of features from CLIPping and HAMLET-FFD. CLIPping (top) forms dataset-specific clusters (FF++, CDF, DFDCP, WDF), reflecting source-dependent representations. In contrast, HAMLET-FFD (bottom) yields clear real/fake clusters (``top-left" vs ``bottom-right") across datasets, indicating domain-invariant authenticity encoding.}
  \label{tsne}
\end{figure}

\subsubsection{\textbf{Emergence of Specialized Detectors}}
\label{Emergence_of_Specialized_Detectors}
Figure~\ref{attention} illustrates distinct attention patterns across embedding types, shedding light on HAMLET-FFD's decision process.
Real embeddings focus on global facial harmony, attending to symmetry and natural feature relationships.
Fake embeddings emphasize regions prone to manipulation, such as eyes, mouth corners, and facial boundaries.
Context embeddings display adaptive behavior: instead of fixed spatial focus, they dynamically shift attention based on image content. The same token (e.g., Context-3) may prioritize central features in one face and symmetry in another.
This emergent behavior forms a distributed ensemble of flexible detectors, each attending to different facial aspects. Such adaptability likely enhances robustness across domains by tailoring analysis to varied forgery styles.
Importantly, this mechanism relies on the visual-to-text pathway. 
Removing this connection hinders adaptive attention allocation, compromising the model’s ability to deploy specialized detectors and weakening its forgery analysis capacity.

\subsubsection{\textbf{Advantages of Bidirectional Cross-Modal Reasoning}}
Figure~\ref{tsne} compares how different CLIP adaptations organize their feature space across datasets. To interpret this, we observe how samples cluster in the top vs. bottom panels.
In CLIPping (top), embeddings group primarily by dataset—real and fake samples from the same source cluster together. This is especially evident in the right-side panels, where each dataset forms a distinct region. Such clustering suggests CLIPping encodes source-specific artifacts rather than universal cues of authenticity, limiting its cross-domain transferability.
HAMLET-FFD, in contrast, clusters embeddings by authenticity: real and fake samples consistently align across datasets. 
Notably, real samples (blue/green) populate a continuous band in the upper-left half of the t-SNE plot, while fake samples (red/purple) are segregated into the lower-right half.
This visualization provides intuitive evidence of HAMLET-FFD’s core strength: its ability to abstract beyond dataset-specific biases by grounding forgery detection in semantically aligned, authenticity-centric representations.

\section{Conclusion}
We proposed HAMLET-FFD, a face forgery detection framework that integrates hierarchical CLIP features through bidirectional cross-modal fusion. 
By enabling mutual refinement between visual representations and textual semantics, our method achieves strong cross-domain generalization without modifying CLIP’s pre-trained parameters.
Extensive experiments across diverse benchmarks demonstrate state-of-the-art performance, including a 90.07\% cross-domain AUC on DeepfakeBench and robust results on emerging manipulation types. 
Ablation studies highlight the critical role of bidirectional fusion, which not only boosts accuracy but also induces specialized, adaptive attention patterns that offer interpretability into the model’s decision process.
As a lightweight plugin atop CLIP, HAMLET-FFD is easily deployable and compatible with broader multimodal systems. 
In future work, we plan to extend our framework to video forgery detection via temporal modeling, and explore broader adaptation strategies to further enhance robustness across modalities and tasks.
\bibliographystyle{ACM-Reference-Format}
\bibliography{hamlet-ffd}

\clearpage
\appendix
\section{Additional Experiments}
\label{sec:appendix}

\subsection{Cross-Manipulation Generalization}
\label{subsec:cross_manipulation}
In accordance with the cross-manipulation evaluation protocol introduced in DeepfakeBench~\cite{DeepfakeBench_YAN_NEURIPS2023}, we further examine HAMLET-FFD’s generalization capability across various manipulation techniques in the FF++ dataset.
This experiment helps understand whether features learned from one forgery method transfer to others—a critical capability for real-world scenarios where the specific manipulation technique is unknown.
For this analysis, we train separate models on each individual manipulation technique (FF-DF, FF-F2F, FF-FS, and FF-NT) and evaluate their performance across all four techniques. 
Table~\ref{tab:cross_manipulation} presents the resulting AUC scores.

Our results reveal several noteworthy patterns. 
Models trained on FF-NT demonstrate superior generalization capability (94.20\% average AUC), suggesting that the textural patterns learned from neural texture manipulations provide valuable cues for detecting other types of forgeries. The NT-trained model performs remarkably well on DF detection (97.76\%), approaching the performance of models specifically trained on DF data.
Conversely, all models struggle when generalizing to FF-NT (performance ranging from 66.33\% to 69.08\%), indicating that neural texture manipulations contain unique characteristics not present in identity-swapping methods. This asymmetric transfer behavior underscores both the capabilities of HAMLET-FFD's hierarchical fusion mechanism in capturing transferable forgery indicators and the inherent differences between manipulation techniques.
Models trained on FF-F2F also show strong cross-technique performance (90.59\% average), particularly when transferring to DF (98.36\%) and FS (96.32\%). This suggests common artifacts may be shared between Face2Face manipulations and other identity-swapping techniques, which HAMLET-FFD effectively captures through its bidirectional feature exchange.

\subsection{Effect of Different CLIP Backbones}
\label{subsec:clip_backbones}
To investigate how the choice of CLIP backbone affects HAMLET-FFD's performance, we evaluate our method using four different CLIP architectures: ResNet-50 (RN50), ResNet-101 (RN101), ViT-Base (ViT-B/16), and ViT-Large (ViT-L/14, our default configuration). 
This analysis helps understand the relationship between model capacity and forgery detection capabilities, providing insights for deployment scenarios with different computational constraints.

When adapting our hierarchical feature extraction approach for ResNet architectures, we extract features from all four stages of the network. 
Since these stages have varying spatial resolutions due to downsampling operations and different channel dimensions, we implement feature alignment techniques. 
Specifically, we use spatial interpolation to unify the spatial dimensions across features from different stages. 
Additionally, we employ 1×1 convolutions for each stage to standardize the channel dimensions, ensuring consistent feature representation before the bidirectional fusion process.

Table~\ref{tab:clip_backbones} presents the AUC scores for both within-domain and cross-domain evaluations across the four backbones.
Our results show a clear correlation between model capacity and forgery detection performance. 
Across all backbones, we observe consistent improvements as model size increases, with the largest gains seen in cross-domain scenarios. The performance gap between RN50 and ViT-L is particularly pronounced for cross-domain evaluation (10.02pp), highlighting the importance of model capacity for generalization to unseen manipulation techniques.
Notably, Vision Transformer architectures substantially outperform their ResNet counterparts of similar parameter counts. 
For example, ViT-B achieves 86.17\% cross-domain AUC compared to 81.79\% for RN101, despite having comparable computational requirements. 
This suggests that the self-attention mechanism in transformer architectures may be particularly well-suited for capturing the subtle inconsistencies present in manipulated faces across different domains.
The superior performance of larger models, especially ViT-L, validates our choice of CLIP:ViT-L/14 as the default backbone for HAMLET-FFD. 
However, the strong results achieved by ViT-B (86.17\% cross-domain AUC) indicate that our method can still deliver competitive performance with more computationally efficient backbones, offering flexibility for resource-constrained deployment scenarios.

\begin{table}[t]\footnotesize
\caption{Cross-manipulation generalization (AUC \%). Models are trained on single manipulation techniques (rows) and evaluated on each technique (columns). Diagonal values (bold) represent within-technique performance.}
\label{tab:cross_manipulation}
\centering
\resizebox{\columnwidth}{!}{
\begin{tabular}{l|>{\columncolor{red!15}}c c c >{\columncolor{red!15}}c|c}
\toprule
\textbf{Training} & \textbf{FF-DF} & \textbf{FF-F2F} & \textbf{FF-FS} & \textbf{FF-NT} & \textbf{Avg.} \\
\midrule
\textbf{FF-DF}    & \textbf{99.39} & 82.48           & 97.60          & 66.33          & 86.45            \\
\textbf{FF-F2F}   & 98.36          & \textbf{98.60}  & 96.32          & 69.08          & 90.59            \\
\textbf{FF-FS}    & 98.47          & 83.40           & \textbf{99.47} & 66.71          & 87.01            \\
\rowcolor{red!10}\textbf{FF-NT}   & 97.76          & 87.18           & 94.87          & \textbf{96.99} & \textbf{94.20}  \\
\bottomrule
\end{tabular}
}
\end{table}

\begin{table}[t]\footnotesize
\caption{Performance comparison (AUC \%) of HAMLET-FFD with different CLIP backbones on within-domain and cross-domain evaluations. Larger models consistently improve performance, with ViT architectures outperforming ResNet counterparts.}
\label{tab:clip_backbones}
\centering
\resizebox{\columnwidth}{!}{
\begin{tabular}{l|cc}
\toprule
\textbf{CLIP Backbone} & \textbf{Within-domain } & \textbf{Cross-domain } \\
\midrule
{RN50}                   & {93.17}                      & {80.05}                     \\
{RN101}                  & {93.44}                      & {81.79}                     \\
{ViT-B/16}                  & {96.34}                      & {86.17}                     \\
\rowcolor{gray!15}{ViT-L/14} & {\textbf{97.73}}             & {\textbf{90.07}}            \\
\bottomrule
\end{tabular}
}
\end{table}

\subsection{Comparison Across Different VLMs}
\label{subsec:different_vlms}

While our main experiments utilize the original CLIP model as the foundation, we extend our analysis to examine HAMLET-FFD's performance when applied to other vision-language models. 
This comparison evaluates whether our bidirectional fusion approach generalizes across different VLM architectures and pre-training strategies. 
We test three representative VLMs: the original CLIP \cite{radford2021learning}, OpenCLIP \cite{ilharco_gabriel_2021_5143773}, and SigLIP \cite{zhai2023sigmoid}, each with both ViT-B and ViT-L backbone variants.

Table~\ref{tab:different_vlms} presents the performance comparison across these VLM variants for both within-domain and cross-domain evaluations.
Our results reveal several interesting findings. 
First, the original CLIP model consistently outperforms other VLMs in cross-domain scenarios, with CLIP:ViT-L achieving the highest cross-domain AUC of 90.07\%. 
This suggests that CLIP's pre-training strategy may develop representations that better capture universal authenticity cues compared to other VLMs. 
Interestingly, for within-domain performance, all three model families achieve comparable results when using the same backbone architecture (differences within 0.7pp for ViT-B and 0.2pp for ViT-L). This indicates that while different pre-training approaches yield similar capabilities for detecting known manipulation techniques, they differ significantly in their ability to generalize to unseen forgeries.
Across all VLM families, the ViT-L backbone consistently outperforms ViT-B, confirming our earlier finding that model capacity plays a crucial role in forgery detection performance. The improvement is particularly pronounced for cross-domain generalization, with gains of 3.90pp, 3.48pp, and 3.44pp for CLIP, OpenCLIP, and SigLIP, respectively.

These results validate our choice of CLIP:ViT-L as the default backbone for HAMLET-FFD. However, they also demonstrate that our bidirectional fusion approach effectively generalizes across different VLM architectures and pre-training strategies, offering flexibility in model selection based on specific application constraints.

\begin{table}[t]\footnotesize
\caption{Performance comparison (AUC \%) of HAMLET-FFD across different vision-language models. For each model family, we test both ViT-B and ViT-L variants to evaluate the impact of model capacity across different architectures.}
\label{tab:different_vlms}
\centering
\resizebox{\columnwidth}{!}{
\begin{tabular}{l|l|cc}
\toprule
\textbf{Model Family} & \textbf{Backbone} & \textbf{Within-domain } & \textbf{Cross-domain } \\
\midrule
\multirow{2}{*}{\small{CLIP}} & ViT-B & \small{96.34} & \small{86.17} \\
                      & \cellcolor{gray!15}ViT-L & \cellcolor{gray!15}\small{\textbf{97.73}} & \cellcolor{gray!15}\small{\textbf{90.07}} \\
\midrule
\multirow{2}{*}{\small{OpenCLIP}} & ViT-B & \small{97.04} & \small{85.26} \\
                          & ViT-L & \small{97.55} & \small{88.74} \\
\midrule
\multirow{2}{*}{\small{SigLIP}} & ViT-B & \small{96.55} & \small{84.94} \\
                        & ViT-L & \small{97.58} & \small{88.38} \\
\bottomrule
\end{tabular}
}
\end{table}

\subsection{Analysis of Attention Heads in HAMLET}
\label{subsec:attention_heads}

The number of attention heads in transformer-based architectures is a critical hyperparameter that affects both model performance and computational efficiency. 
To determine the optimal configuration for the HAMLET module's multi-head attention components, we conduct a systematic study varying the number of attention heads from 4 to 128, while keeping all other hyperparameters constant.

\begin{table}[t]\footnotesize
\caption{Impact of attention head count on HAMLET-FFD performance (AUC \%). Performance initially improves with more attention heads, peaking at 32 heads, before declining slightly with further increases.}
\label{tab:attention_heads}
\centering
\resizebox{\columnwidth}{!}{
\begin{tabular}{l|cc}
\toprule
\textbf{Number of Heads} & \textbf{Within-domain } & \textbf{Cross-domain } \\
\midrule
4                       & 97.01                      & 89.12                     \\
8                       & 97.13                      & 89.28                     \\
16                      & 97.56                      & 89.67                     \\
\rowcolor{gray!15}32 (default) & \textbf{97.73}             & \textbf{90.07}            \\
64                      & 97.67                      & 89.99                     \\
128                     & 97.42                      & 89.91                     \\
\bottomrule
\end{tabular}
}
\end{table}

Table~\ref{tab:attention_heads} presents the AUC scores for both within-domain and cross-domain evaluations across the different attention head configurations.
Our results reveal a clear trend of initially improving performance as the number of attention heads increases, followed by diminishing returns and eventually slight performance degradation with too many heads. Performance peaks at 32 attention heads for both within-domain (97.73\% AUC) and cross-domain (90.07\% AUC) evaluations.
This pattern aligns with theoretical understanding of multi-head attention mechanisms. With too few heads, the model lacks the capacity to simultaneously attend to different aspects of the input. In our case, forgery detection requires attention to various facial regions and feature types simultaneously, explaining why configurations with fewer heads (4, 8) yield suboptimal results.
Conversely, configurations with too many heads (64, 128) show slight performance degradation. This may occur because each head receives a smaller embedding dimension as the number of heads increases, potentially limiting the expressiveness of individual attention patterns. The degradation is more pronounced for within-domain performance, suggesting that excessive attention fragmentation may particularly affect the model's ability to capture fine-grained patterns in familiar manipulation techniques.

Based on these findings, we selected 32 attention heads as the default configuration for HAMLET-FFD, providing the optimal balance between model expressiveness and computational efficiency. This configuration allows sufficient parallel attention to different facial regions while maintaining adequate representational capacity within each attention head.

\subsection{Impact of Weight Coefficients in the Training Objective}
\label{subsec:weight_coefficients}

Our training objective employs a progressive cross-entropy loss with decreasing weight coefficients to enforce a similarity hierarchy. The default configuration uses geometric decay with a factor of 0.5, expressed as $(0.5)^i$ for position $i$ in the similarity sequence. To understand the sensitivity of our method to these coefficients, we experiment with various weighting schemes while maintaining the same hierarchical similarity structure.

\begin{table}[t]\small
\caption{Performance comparison (AUC \%) of HAMLET-FFD with different weight coefficient schemes in the training objective. The geometric decay weighting scheme (default) achieves the best performance.}
\label{tab:weight_coefficients}
\centering
\resizebox{\columnwidth}{!}{
\begin{tabular}{l|cc|l}
\toprule
\textbf{Weighting Scheme} & \textbf{Within-domain} & \textbf{Cross-domain} & \textbf{Coefficient Formula} \\
\midrule
\rowcolor{gray!15}Geometric decay & \normalsize{\textbf{97.73}} & \normalsize{\textbf{90.07}} & $w_i = (0.5)^i,\ i=0,1,2$ \\
Uniform weighting & \normalsize{97.19} & \normalsize{88.94} & $w_i = 1,\ i=0,1,2$ \\
Linear decay      & \normalsize{97.52} & \normalsize{89.66} & $w_i = 1-\frac{i}{3},\ i=0,1,2$ \\
Harmonic decay    & \normalsize{97.69} & \normalsize{89.84} & $w_i = \frac{1}{i+1},\ i=0,1,2$ \\
Cosine decay      & \normalsize{97.47} & \normalsize{89.17} & $w_i = \cos(\frac{i\pi}{2N}),\ i=0,1,2;\ N=3$ \\
\bottomrule
\end{tabular}
}
\end{table}

\begin{figure*}[t] 
  \centering
  \includegraphics[width=\linewidth]{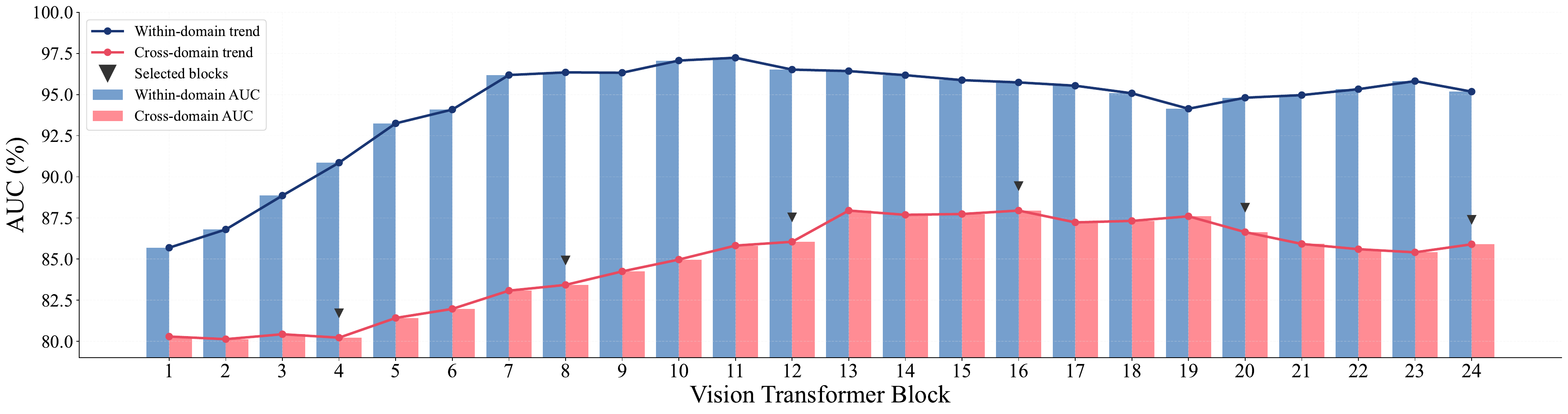}
  \caption{Performance analysis of individual transformer blocks. The light blue bars and dark blue trend line show within-domain performance, which peaks at blocks 10-11. The light pink bars and red trend line represent cross-domain generalization, which reaches maximum at blocks 13-16. Black triangles mark the blocks selected for our multi-level feature fusion approach (blocks 4, 8, 12, 16, 20, and 24).}
\label{fig:single_block_performance}
\end{figure*}

Table~\ref{tab:weight_coefficients} presents the AUC scores for both within-domain and cross-domain evaluations across different weighting schemes.
Our results show that the weighting scheme significantly impacts model performance, particularly for cross-domain generalization. The geometric decay scheme (our default configuration) achieves the best performance with 90.07\% cross-domain AUC, outperforming all other weighting strategies. This suggests that a rapidly decreasing emphasis on lower-ranked similarities in the hierarchy is beneficial for learning robust authenticity representations.
Uniform weighting, which treats all similarity levels equally, yields the lowest performance (88.94\% cross-domain AUC), representing a drop of 1.13pp compared to the geometric decay scheme. This confirms the importance of the progressive weighting approach in our hierarchical similarity objective.
Among the alternative decay schemes, harmonic decay performs closest to the geometric decay (89.84\% cross-domain AUC), followed by linear decay (89.66\%) and cosine decay (89.17\%). The relatively strong performance of harmonic decay may be attributed to its balance between rapid initial decay and sustained emphasis on later terms in the sequence.

These findings validate our choice of geometric decay for the default configuration of HAMLET-FFD, demonstrating that the specific progression of weights in the similarity hierarchy plays a meaningful role in developing robust cross-domain representations for face forgery detection.

\subsection{Single-Block vs. Multi-Level Feature Analysis}
\label{subsec:single_block}
A core innovation of HAMLET-FFD is its use of hierarchical features from multiple transformer blocks. 
To quantify the contribution of each individual block and validate the benefits of our multi-level approach, we conduct experiments using patch embeddings from each block separately, revealing how different layers of CLIP's vision transformer capture forgery-relevant information.

Figure~\ref{fig:single_block_performance} presents the AUC scores for both within-domain and cross-domain evaluations when using features from each individual transformer block.
Our analysis reveals distinct patterns in how different transformer blocks contribute to face forgery detection. 
For within-domain evaluation, performance follows a clear arc: it starts relatively low (85.69\% at block 1), increases steadily through the earlier layers, and peaks at blocks 10-11 (97.07\%-97.24\%). 
After reaching this maximum, performance moderately decreases in the middle layers before stabilizing in the deeper layers, ending at 95.18\% for block 24.
For cross-domain generalization, we observe a different trajectory. 
Performance begins at around 80\% for the earliest blocks and shows modest improvement until approximately block 12 (86.05\%).
From blocks 13-19, performance plateaus at its highest levels (87.23\%-87.95\%), with a slight peak at blocks 13 and 16 (both 87.95\%). 
This is followed by a gradual decline in the final layers.
The key distinction between these patterns is their phase shift: within-domain performance peaks at middle blocks (10-11), while cross-domain performance reaches its maximum at deeper blocks (13-16). 
This phase difference suggests that middle blocks excel at capturing dataset-specific artifacts, while slightly deeper blocks develop more generalizable representations that transfer better to unseen manipulation techniques.

Based on this analysis, our multi-level approach strategically samples features from blocks distributed throughout the network (blocks 4, 8, 12, 16, 20, and 24). 
This selection spans the range from early feature extraction to deep semantic representation, allowing HAMLET-FFD to leverage complementary information across different abstraction levels. The result is significantly improved performance (97.73\% within-domain and 90.07\% cross-domain AUC) compared to any single block.
The substantial performance gap between our multi-level approach and even the best individual block (87.95\% for cross-domain) validates the effectiveness of our hierarchical feature fusion mechanism. 
This improvement demonstrates that integrating information from multiple levels of abstraction is essential for robust cross-domain forgery detection.

\section{Additional Evaluation Metrics}
\label{subsec:additional_metrics}
While our main paper primarily focuses on frame-level AUC as the evaluation metric, we provide additional performance measurements here to offer a more comprehensive assessment of HAMLET-FFD. Table~\ref{tab:additional_metrics} presents results across multiple metrics: accuracy (ACC), average precision (AP), equal error rate (EER), and video-level AUC, alongside the previously reported frame-level AUC.

Consistent with our frame-level AUC results, HAMLET-FFD demonstrates strong performance across all metrics. Video-level AUC consistently exceeds frame-level AUC (by 1.17-4.21 percentage points on average), confirming that aggregating predictions across frames enhances detection reliability. This improvement is particularly pronounced for challenging datasets like DFDC (85.39\% to 88.04\%) and BlendFace (89.43\% to 92.75\%).
The Equal Error Rate (EER) results further validate our method's effectiveness, with notably low values for DiffSwap (3.15\%) and UADFV (4.33\%). The higher EER values on datasets like DFDC (23.16\%) and Fsh (25.59\%) correlate with their known challenges in cross-domain generalization.
Average Precision (AP) scores are consistently high, exceeding 90\% on most datasets and reaching 99.41\% on FF++ and 99.31\% on DFD. This indicates HAMLET-FFD's strong capability to maintain precision across different recall thresholds, an important consideration for practical deployment scenarios where false positive rates must be minimized.
\begin{table*}[ht]
    \caption{Performance of HAMLET-FFD across multiple evaluation metrics. Results are shown for all datasets used in our experiments, including DeepfakeBench datasets and extended benchmarks. All values are percentages except for EER (lower is better). Video-level evaluation is not applicable for DiffSwap due to the dataset structure.}
    \label{tab:additional_metrics}
    \centering
    \resizebox{\textwidth}{!}{
    \begin{tabular}{l|c|ccccccc|c|c|c|cccc}
    \toprule
    \multirow{2}{*}{\textbf{Metric}} & \multirow{2}{*}{\textbf{FF++}} & \multicolumn{7}{c|}{\textbf{DeepfakeBench}} & \multirow{2}{*}{\textbf{WDF}} & \multirow{2}{*}{\textbf{FFIW}} & \multirow{2}{*}{\textbf{DiffSwap}} & \multicolumn{4}{c}{\textbf{DF40}}  \\
    \cmidrule(lr){3-9} \cmidrule(lr){13-16}
    ~ & ~ & \textbf{CDFv1} & \textbf{CDFv2} & \textbf{DFD} & \textbf{DFDC} & \textbf{DFDCP} & \textbf{Fsh} & \textbf{UADFV} & ~ & ~ & ~ & \textbf{UniFace} & \textbf{E4S} & \textbf{BlendFace} & \textbf{MobileSwap} \\
    \midrule
    AUC (frame) & 97.68 & 90.46 & 89.46 & 94.35 & 85.39 & 89.60 & 82.29 & 98.95 & 88.44 & 86.60 & 99.03 & 91.99 & 95.71 & 89.43 & 95.21 \\
    ACC         & 93.75 & 81.92 & 81.52 & 89.75 & 76.25 & 82.13 & 72.88 & 94.94 & 73.31 & 75.17 & 90.75 & 81.27 & 88.74 & 77.58 & 84.05 \\
    AP          & 99.41 & 94.15 & 94.21 & 99.31 & 87.90 & 94.21 & 83.26 & 99.22 & 87.95 & 86.94 & 98.44 & 91.07 & 97.27 & 88.46 & 98.98 \\
    EER         & 7.14  & 17.54 & 18.84 & 13.06 & 23.16 & 19.14 & 25.59 & 4.33  & 19.76 & 22.35 & 3.15 & 15.98  & 11.15 & 18.46 & 11.34 \\
    \midrule
    AUC (video) & 98.85 & 91.43 & 93.67 & 97.76 & 88.04 & 92.13 & 84.53 & 99.42 & 90.53 & 88.66 & --- & 94.99 & 98.73 & 92.75 & 97.97 \\
    \bottomrule
    \end{tabular}
    }
\end{table*}

\section{Additional Attention Visualization}
\label{subsec:attention_vis}
\begin{figure*}[h]
\centering
\vspace{20mm}
\includegraphics[width=\linewidth]{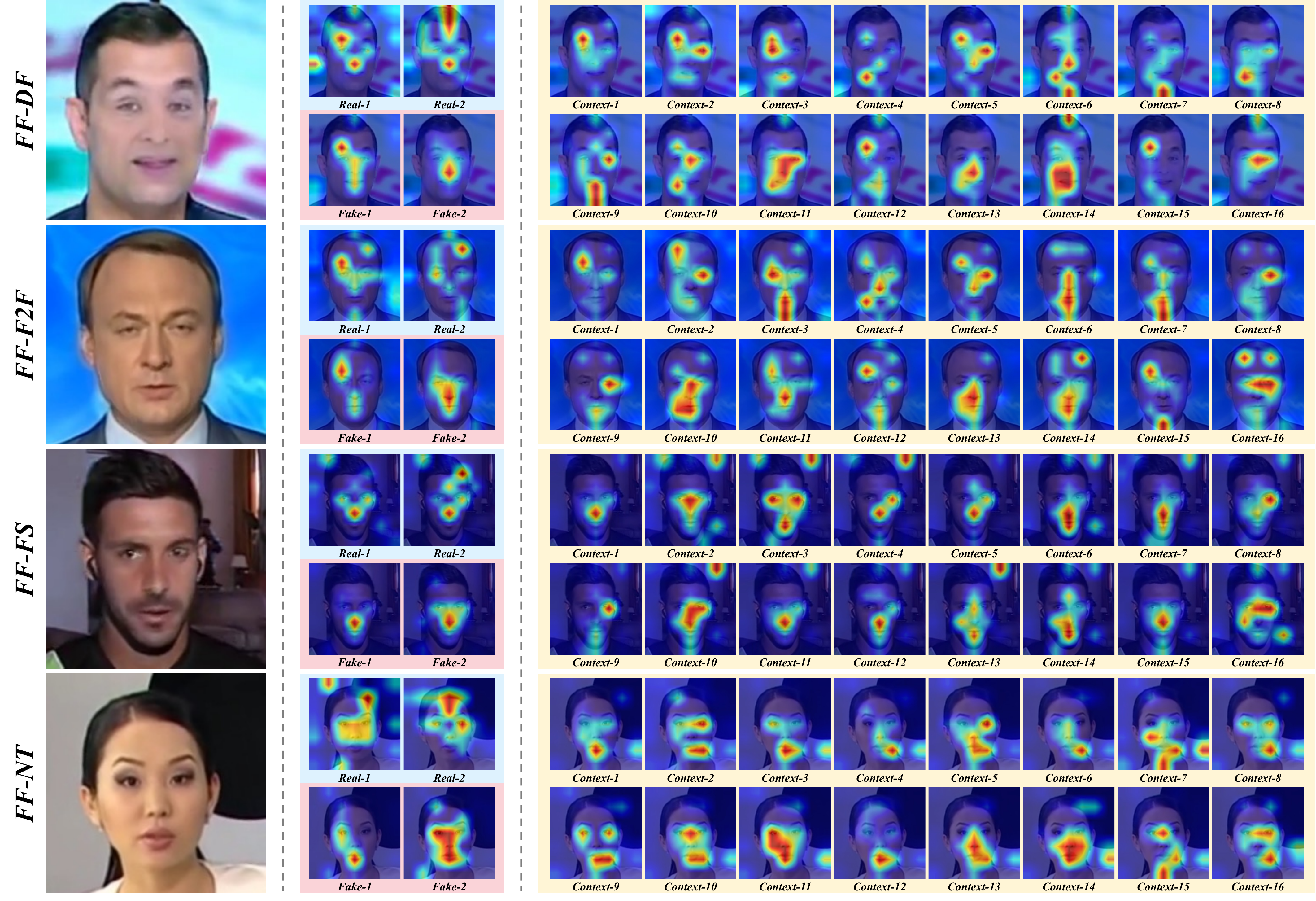}
\vspace{-0.25in}
\caption{Attention patterns on FF++ manipulations (FF-DF, FF-F2F, FF-FS, FF-NT).}
\label{fig:attention_ff}
\end{figure*}

\begin{figure*}[t]
\centering
\includegraphics[width=\linewidth]{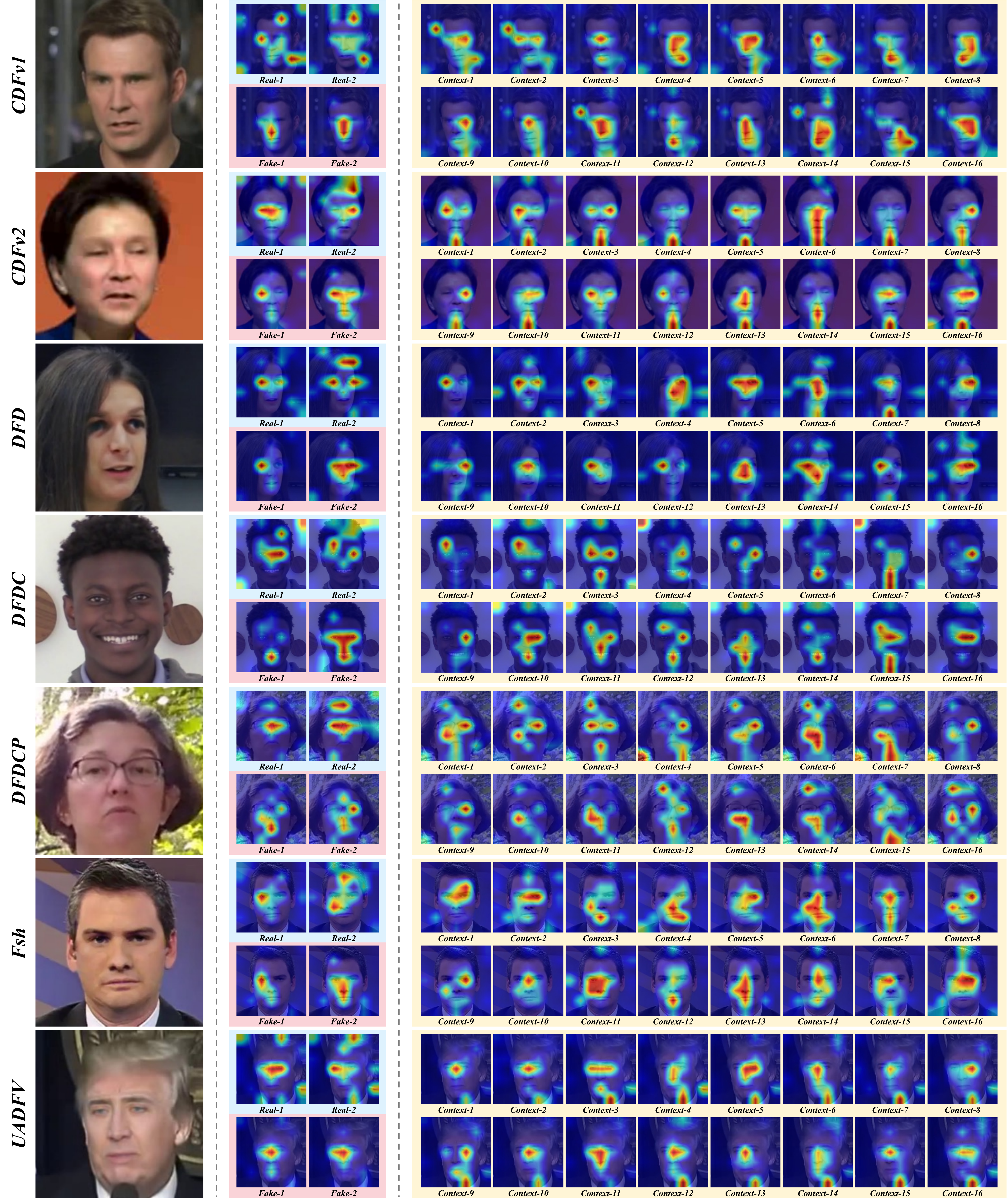}
\vspace{-0.25in}
\caption{Attention patterns on DeepfakeBench cross-domain datasets (CDFv1, CDFv2, DFD, DFDC, DFDCP, Fsh, UADFV).}
\label{fig:attention_dfb}
\end{figure*}

\begin{figure*}[t]
\centering
\includegraphics[width=\linewidth]{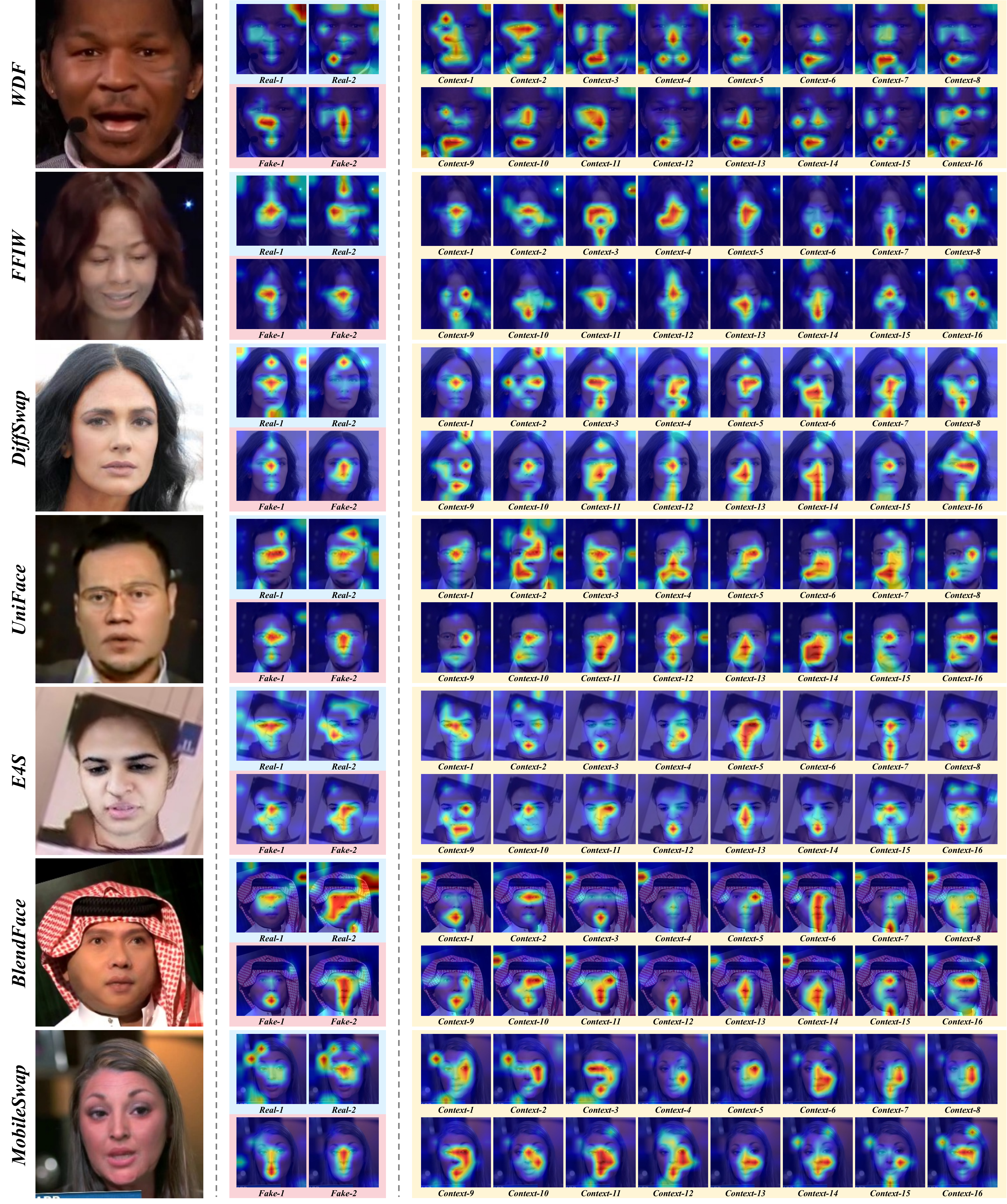}
\vspace{-0.25in}
\caption{Attention patterns on recent and emerging forgery techniques: in-the-wild forgeries (WDF, FFIW), diffusion-based manipulations (DiffSwap), and DF40 subsets (UniFace, E4S, BlendFace, MobileSwap).}
\label{fig:attention_extended}
\end{figure*}

To provide deeper insights into how HAMLET-FFD analyzes different types of facial manipulations, we present extended attention visualization results across various datasets (Figure~\ref{fig:attention_ff}, \ref{fig:attention_dfb}, \ref{fig:attention_extended}). 
These visualizations expand upon the Section~\ref{Emergence_of_Specialized_Detectors} \textit{``Emergence of Specialized Detectors"} analysis.
These extended visualizations across diverse datasets complement our quantitative results, providing intuitive evidence for how the bidirectional fusion mechanism enables specialized attention patterns that adapt to various manipulation techniques while maintaining consistent forensic reasoning.

\section{Implementation Details}
\label{subsec:implementation}
To facilitate reproducibility and provide insights into our implementation, we present the core algorithm of HAMLET-FFD's bidirectional fusion mechanism. 
The implementation follows the architecture described in Section~\ref{sec/3_methodology}, with particular focus on the hierarchical feature access and bidirectional modal fusion components.

The core of our approach is the Hierarchical Modal Fusion module, which implements the bidirectional conditioning between visual and textual modalities. Code~\ref{code:hierarchical_fusion} shows the PyTorch implementation of this module.
The bidirectional fusion process operates in two primary stages as illustrated in the code. 
First, the text-to-visual conditioning pathway modulates multi-level visual features using parameters derived from textual embeddings. 
This enables the model to interpret visual information through the lens of specialized authenticity concepts. We implement this as feature-wise affine transformations applied to patch embeddings from multiple transformer blocks.
Second, the visual-to-text conditioning pathway enables visual evidence to refine textual concepts through cross-attention.
This creates a closed feedback loop where textual embeddings guide visual feature interpretation, and aggregated visual features subsequently update these embeddings with image-specific information.

The main HAMLET module (shown in Code~\ref{code:hamlet}) orchestrates the overall process, managing the specialized authenticity embeddings and coordinating the bidirectional fusion.
A key implementation detail is our hierarchical feature integration approach. Visual features from six strategically selected transformer blocks are extracted using forward hooks and then processed through self-attention to create level-aware representations. These are combined through average pooling to produce comprehensive, multi-scale feature maps before the cross-attention operation.

During training, we employ a progressive cross-entropy loss with geometric decay weights ($(0.5)^i$) to enforce the similarity hierarchy described in Section~\ref{sec/3_methodology}. Our implementation preserves all CLIP parameters in their original, frozen state by accessing intermediate features through hooks without modifying the original network. This design choice enables HAMLET-FFD to function as an external plugin that enhances CLIP's capabilities for forgery detection without compromising its utility for other vision-language tasks.

\setcounter{figure}{0} 
\definecolor{codegray}{gray}{0.15}
\definecolor{codebg}{rgb}{0.95,0.95,0.95}
\definecolor{darkgreen}{RGB}{0,100,0}

\begin{figure*}[ht]
  \centering
  \label{code:hierarchical_fusion}
  \captionsetup{labelformat=empty} 
  \caption{PyTorch implementation of the Hierarchical Modal Fusion module}
    \begin{lstlisting}[language=Python, 
                       label=code:hierarchical_fusion,
                       basicstyle=\footnotesize\ttfamily,  % 或 \small
                      keywordstyle=\color{blue!70!black}\bfseries,
                      commentstyle=\color{darkgreen}\itshape,
                      stringstyle=\color{orange},
                      numberstyle=\tiny\color{gray},
                      breaklines=true,
                      showstringspaces=false]
class HierarchicalModalFusion(nn.Module):
    def __init__(self, text_embed_dim, visual_embed_dim, num_heads):
        super().__init__()
        self.text_embed_dim = text_embed_dim
        self.visual_embed_dim = visual_embed_dim
        self.num_heads = num_heads

        # Setup for bidirectional fusion
        self.mlp = nn.Sequential(
            nn.SiLU(),
            nn.Linear(text_embed_dim, visual_embed_dim * 2)
        )

        self.ln_self_attn = nn.LayerNorm(visual_embed_dim)
        self.self_attention = nn.MultiheadAttention(
            embed_dim=visual_embed_dim,
            num_heads=num_heads,
            batch_first=True,
        )

        # Normalization layers for text (query) and visual (key/value) embeddings
        self.ln_q = nn.LayerNorm(text_embed_dim)
        self.ln_k = nn.LayerNorm(visual_embed_dim)
        self.cross_attention = nn.MultiheadAttention(
            embed_dim=text_embed_dim,  # Dimension of the query (text embedding space)
            num_heads=num_heads,       # Number of attention heads
            batch_first=True,          # Input tensors have batch as the first dimension
            kdim=visual_embed_dim,     # Dimension of the key embeddings (visual embedding space)
            vdim=visual_embed_dim      # Dimension of the value embeddings (visual embedding space)
        )

    def forward(self, text_emb: torch.Tensor, multi_layer_patch_emb: torch.Tensor):
        num_layers, batch_size, num_patches, d = multi_layer_patch_emb.shape
        multi_layer_patch_emb = multi_layer_patch_emb.permute(1, 2, 0, 3).flatten(0, 1)  # n b l d -> (bl) n d

        # First stage: text -> visual conditioning
        scale, shift = (self.mlp(text_emb.mean(dim=1))[:, None, None, :].expand(-1, num_patches, num_layers, -1).flatten(0, 1)).chunk(2, dim=-1)
        conditioned_patch_emb = multi_layer_patch_emb * (scale + 1) + shift  

        # Self attention and residual connection
        self_attn_input = self.ln_self_attn(conditioned_patch_emb)
        fused_patch_emb = (multi_layer_patch_emb + self.self_attention(self_attn_input, self_attn_input, self_attn_input)[0]).mean(dim=1).view(batch_size, num_patches, -1)

        # Second stage: visual -> text conditioning
        query = self.ln_q(text_emb)
        key = self.ln_k(fused_patch_emb)
        conditioned_text_emb = text_emb + self.cross_attention(query, key, key)[0]

        return conditioned_text_emb
         
\end{lstlisting}
\end{figure*}

\begin{figure*}[ht]
  \centering
  \label{code:hamlet}
  \captionsetup{labelformat=empty} 
  \caption{Core implementation of the HAMLET module}
\begin{lstlisting}[language=Python, 
                   label=code:hamlet,
                   basicstyle=\footnotesize\ttfamily,  % 或 \small
                  keywordstyle=\color{blue!70!black}\bfseries,
                  commentstyle=\color{darkgreen}\itshape,
                  stringstyle=\color{orange},
                  numberstyle=\tiny\color{gray},
                  breaklines=true,
                  showstringspaces=false]
class HAMLET(nn.Module):
    def __init__(self, clip_model, num_authenticity_tokens, num_context_tokens, num_heads, dropout_rate=0.1):
        super().__init__()
        self.num_heads = num_heads
        # Initialize learnable authenticity embeddings
        assert num_authenticity_tokens + num_context_tokens + 2 <= clip_model.context_length
        text_embed_dim = clip_model.token_embedding.embedding_dim
        scale = text_embed_dim ** -0.5
        self.real_embeddings = nn.Parameter(scale * torch.randn(1, num_authenticity_tokens, text_embed_dim))
        self.fake_embeddings = nn.Parameter(scale * torch.randn(1, num_authenticity_tokens, text_embed_dim))
        self.context_embeddings = nn.Parameter(scale * torch.randn(1, num_context_tokens, text_embed_dim))
        # Dropout layers
        self.real_embed_dropout = nn.Dropout(dropout_rate)
        self.fake_embed_dropout = nn.Dropout(dropout_rate)
        self.context_embed_dropout = nn.Dropout(dropout_rate)

    def forward(self, image_tensor: torch.Tensor):
        with torch.no_grad():
            image_features = self.__encode_image(image_tensor)

        # get multi_level_patch_embeddings
        multi_layer_patch_emb = torch.stack(self.multi_layer_visual_embeddings)  # [n, b, l, d]
        batch_size = multi_layer_patch_emb.size(1)
        self.reset_hierarchical_features()  # clear the cache

        # get conditioned authenticity embeddings
        context_embeddings = self.cross_modal_fusion_module(self.context_embed_dropout(self.context_embeds.repeat(batch_size, 1, 1)), multi_layer_patch_emb)
        real_embeddings = self.cross_modal_fusion_module(self.real_embed_dropout(self.real_embeds.repeat(batch_size, 1, 1)), multi_layer_patch_emb)
        fake_embeddings = self.cross_modal_fusion_module(self.fake_embed_dropout(self.fake_embeds.repeat(batch_size, 1, 1)), multi_layer_patch_emb)
        
        return {
            "image_features": image_features,
            "real_prompt_features": self.__encode_text_embed(torch.cat([real_embeddings, context_embeddings], dim=1)),
            "fake_prompt_features": self.__encode_text_embed(torch.cat([fake_embeddings, context_embeddings], dim=1)),
            "context_features": self.__encode_text_embed(context_embeddings)
        }

    def _bind_encoder_functions(self, clip_model):
        # bind encode image function
        self.__encode_image = lambda x: clip_model.encode_image(x)
        
        sot_embed = clip_model.token_embedding(clip.tokenize("")[0, 0])[None, None, :]  # [1, 1, d]
        eot_embed = clip_model.token_embedding(clip.tokenize("")[0, 1])[None, None, :]  # [1, 1, d]
        pad_embed = clip_model.token_embedding(clip.tokenize("")[0, 2])[None, None, :]  # [1, 1, d]

        def encode_text_embed(text_embed):
            device = text_embed.device
            batch_size = text_embed.size(0)
            text_length = text_embed.size(1)
            x = torch.cat([sot_embed.expand(batch_size, -1, -1).to(device),
                           text_embed,
                           eot_embed.expand(batch_size, -1, -1).to(device),
                           pad_embed.expand(batch_size, clip_model.context_length - 2 - text_length, -1).to(device)],
                          dim=1)
            x = x.type(clip_model.dtype) + clip_model.positional_embedding.type(clip_model.dtype)
            x = x.permute(1, 0, 2)  # n l d -> l n d
            x = clip_model.transformer(x)
            x = x.permute(1, 0, 2)  # l n d -> n l d
            x = clip_model.ln_final(x).type(clip_model.dtype)
            x = x[torch.arange(x.shape[0]), text_length + 1] @ clip_model.text_projection
            return x

        # bind encode text function
        self.__encode_text_embed = encode_text_embed

   
\end{lstlisting}
\end{figure*}

\begin{figure*}[ht]
  \centering
  \captionsetup{labelformat=empty} 
\begin{lstlisting}[language=Python, 
                   basicstyle=\footnotesize\ttfamily,  % 或 \small
                  keywordstyle=\color{blue!70!black}\bfseries,
                  commentstyle=\color{darkgreen}\itshape,
                  stringstyle=\color{orange},
                  numberstyle=\tiny\color{gray},
                  breaklines=true,
                  showstringspaces=false]
    def _bind_vision_language_backbone(self, clip_model, patch_embedding_every_n_block=4):
        self.__logit_scale = clip_model.logit_scale
        self.__hooks = []
        self._bind_encoder_functions(clip_model)
        self.multi_layer_visual_embeddings = []

        self.cross_modal_fusion_module = HierarchicalModalFusion(clip_model.token_embedding.embedding_dim, clip_model.visual.transformer.width, self.num_heads)

        def hook_fn(module, input, output):
            patch_embeddings = output[1:].transpose(0, 1) if not module.attn.batch_first else output[:, 1:]
            self.multi_layer_visual_embeddings.append(patch_embeddings.detach().clone())

        for i in range(len(clip_model.visual.transformer.resblocks)):
            if (i + 1) % patch_embedding_every_n_block == 0:
                handle = clip_model.visual.transformer.resblocks[i].register_forward_hook(hook_fn)
                self.__hooks.append(handle)

    def _unbind_vision_language_backbone(self):
        # unbind hooks
        for handle in self.__hooks:
            handle.remove()
        del self.__hooks

        # unbind multi_layer_visual_embeddings
        self.reset_hierarchical_features()
        del self.multi_layer_visual_embeddings

        # unbind clip related functions
        del self.__logit_scale
        del self.__encode_image
        del self.__encode_text_embed

    def reset_hierarchical_features(self):
        if not hasattr(self, "multi_layer_visual_embeddings"):
            self.multi_layer_visual_embeddings = []
        else:
            self.multi_layer_visual_embeddings.clear()
\end{lstlisting}
\end{figure*}

\end{document}